\documentclass[lettersize,journal]{IEEEtran}
\usepackage{amsmath,amsfonts}
\usepackage{array}
\usepackage{textcomp}
\usepackage{stfloats}
\usepackage{url}
\usepackage{verbatim}
\usepackage{graphicx}
\usepackage{cite}
\usepackage{amsmath}
\usepackage{cite}
\usepackage{bm}
\usepackage{array}  
\usepackage{comment}
\usepackage{threeparttable,multirow,booktabs}
\usepackage[table,xcdraw]{xcolor}
\newcommand{\Tianruinew}[1]{{\color{black}{#1}}}

\definecolor{mygray}{gray}{0.85}
\usepackage{bbding}
\usepackage{textcomp}
\usepackage{url,hyperref,bookmark}
\usepackage{pifont}
\usepackage{amssymb}
\usepackage{xspace}
\usepackage{graphicx}
\usepackage{subfigure}

\usepackage{array}
\newcolumntype{C}[1]{>{\let\newline\\\arraybackslash\hspace{0pt}}m{#1}}
\newcolumntype{L}[1]{>{\raggedright\let\newline\\\arraybackslash\hspace{0pt}}m{#1}}

\begin{document}

\title{SMILENet: Unleashing Extra-Large Capacity Image Steganography via a Synergistic Mosaic InvertibLE Hiding Network}

\author{Jun-Jie~Huang$^{\dagger}$,~\IEEEmembership{Member,~IEEE,}
        Zihan~Chen$^{\dagger}$,
        Tianrui~Liu$^{*}$,
        Wentao~Zhao$^{*}$,
        Xin~Deng,~\IEEEmembership{Member,~IEEE,}
        Xinwang~Liu,~\IEEEmembership{Senior Member,~IEEE,}
        Meng~Wang,~\IEEEmembership{Fellow,~IEEE,}
        and~Pier Luigi~Dragotti,~\IEEEmembership{Fellow,~IEEE}

\IEEEcompsocitemizethanks{\IEEEcompsocthanksitem J.-J. Huang, Z. Chen, T. Liu, W. Zhao, X. Liu are affiliated with the College of Computer Science and Technology, National University of Defense Technology, Changsha, China. E-mail: \{jjhuang, chenzihan21, trliu, wtzhao, xinwangliu\}@nudt.edu.cn.
\IEEEcompsocthanksitem X. Deng is affiliated with the School of Electronic Information Engineering, Beihang University, Beijing, China. E-mail: cindydeng@buaa.edu.cn.
\IEEEcompsocthanksitem M. Wang is affiliated with the School of Computer Science and Information Engineering, Hefei University of Technology, Hefei, China. E-mail: eric.mengwang@gmail.com.
\IEEEcompsocthanksitem P.L. Dragotti is affiliated with the Department of Electrical and Electronic Engineering, Imperial College London, London, UK. E-mail: p.dragotti@imperial.ac.uk.}

\thanks{$^{\dagger}$These authors contributed equally to this work. $^{*}$Corresponding author.}
}

\markboth{Submitted to IEEE Transactions on Pattern Analysis and Machine Intelligence}%
{Shell \MakeLowercase{\textit{et al.}}: A Sample Article Using IEEEtran.cls for IEEE Journals}


\maketitle

\begin{abstract}
Existing image steganography methods face fundamental limitations in hiding capacity (typically $1\sim7$ images) due to severe information interference and uncoordinated capacity-distortion trade-off. We propose SMILENet, a novel synergistic framework that achieves 25 image hiding through three key innovations: (i) A synergistic network architecture coordinates reversible and non-reversible operations to efficiently exploit information redundancy in both secret and cover images. The reversible Invertible Cover-Driven Mosaic (ICDM) module and Invertible Mosaic Secret Embedding (IMSE) module establish cover-guided mosaic transformations and representation embedding with mathematically guaranteed invertibility for distortion-free embedding. The non-reversible Secret Information Selection (SIS) module and Secret Detail Enhancement (SDE) module implement learnable feature modulation for critical information selection and enhancement. (ii) A unified training strategy that coordinates complementary modules to achieve 3.0× higher capacity than existing methods with superior visual quality. (iii) Last but not least, we introduce a new metric to model Capacity-Distortion Trade-off for evaluating the image steganography algorithms that jointly considers hiding capacity and distortion, and provides a unified evaluation approach for accessing results with different number of secret image. Extensive experiments on DIV2K, Paris StreetView and ImageNet1K show that SMILENet outperforms state-of-the-art methods in terms of hiding capacity, recovery quality as well as security against steganalysis methods.
\end{abstract}


\section{Introduction}
Steganography~\cite{anderson1998limits, johnson1998exploring, provos2003hide} is a classical problem which has been extensively studied in the fields of computer vision and signal processing.
By exploiting the inherent redundancy of the cover media, steganography aims to hide secret information in the form of text, audio, image, etc, into a non-secret cover, so that only the intended recipient can recover it.
In particular, image steganography~\cite{lin1999review,cheddad2010digital,hussain2018image,Kadhim2019ComprehensiveSO} hides secret messages into a cover image, striving to maintain visual similarity to the original while ensuring adequate information retrieval from the resulting stego image. Image steganography has a wide range of applications, including covert communication, information protection and digital watermarking. 
From a practical perspective, 
it is crucial to hide as much information as possible imperceptibly while ensuring its safe recovery. Therefore, the literature of image steganography has focused on both enhancing the hiding capacity of the cover image and improving the quality of the recovered secret message.

\begin{figure}[t]
    \centering
    \includegraphics[width=0.95\linewidth]{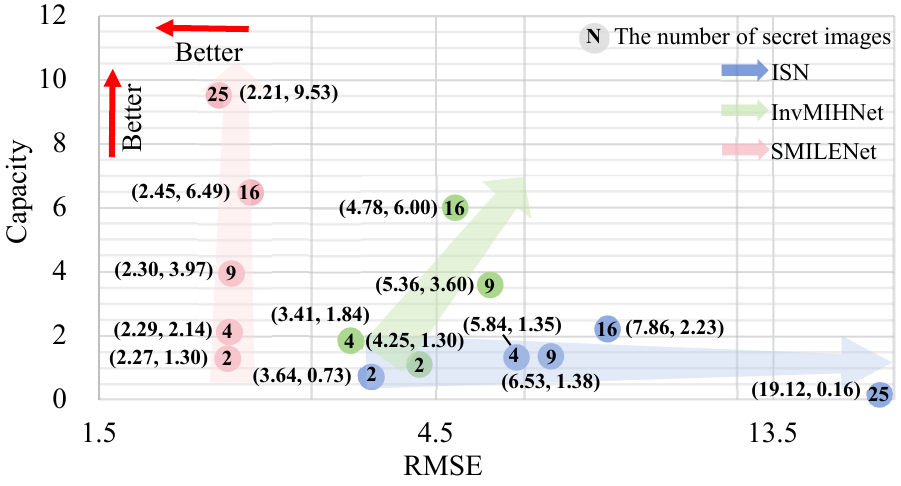}    
    \caption{The Capacity-Distortion Trade-off of hiding different numbers of secret images using different steganography methods, including ISN~\cite{Lu2021LargecapacityIS}, InvMIHNet~\cite{chen2024InvMIHNet}, and SMILENet. The horizontal axis of Capacity-Distortion curve is RMSE of cover/stego image pairs; the vertical axis indicates the hiding capacity of secret images evaluated in terms of the sum of mutual information between input and recovered secret images (see Eqn.~\eqref{eq:CD_fun} for more details). The results are evaluated on DIV2K~\cite{agustsson2017ntire} dataset.}
    \label{fig:capacity_rmse_SMILENet}
    \vspace{-3pt}
\end{figure}

Traditional image steganography methods~\cite{Chan2004HidingDI, nguyen2016adaptive, yang2009high, tseng2014high,barni2001improved,provos2003hide,atawneh2017secure} start to attempt hiding secret messages, typically text, into a cover image by leveraging the inherent information redundancy of the cover image. These methods hide information either in the image domain or the transform domain using handcrafted features and can typically achieve a hiding capacity of $0.2 \sim 4$ bits per pixel (bpp), since they generally require perfect secret information decoding. 

Learning-based methods leverage the strong learning capabilities of deep networks to enable 
image-to-image hiding and recovery, while relaxing the requirement for perfect reconstruction.
The deep networks employed for image steganography can be broadly categorized into two groups: Convolutional Neural Networks (CNNs)-based approaches~\cite{Baluja2017HidingII,baluja2019hiding,rahim2018end,Zhu2018HiDDeNHD,tang2019cnn,zhang2020udh,chen2022hiding}, and Invertible Neural Networks (INNs)-based approaches~\cite{Jing2021HiNetDI,Lu2021LargecapacityIS,xu2022robust,guan2022deepmih,chen2024InvMIHNet}. 
Compared to traditional methods, the learning-based methods have significantly enhanced the hiding capacity and can now successfully hide and recover $1 \sim 7$ full color images within a single cover image. 
However, as more secret images are concealed, these methods often suffer from perceptible color distortion in the recovered secret images due to an increasingly intensified information interference among multiple secret images. 

Fig.~\ref{fig:capacity_rmse_SMILENet} exposes the capacity-distortion trade-off phenomenon in image steganography methods, for example for ISN~\cite{Lu2021LargecapacityIS} and InvMIHNet~\cite{chen2024InvMIHNet} methods.
Here, the hiding capacity is defined as the sum of normalized mutual information between the input and the recovered secret images, and the distortion is measured by RMSE between the cover image and the stego image. 
As we can see, as the number of secret images increases, the results of ISN exhibit an ineffective capacity-distortion trade-off characterized by: (i) a dramatically growing distortion, and (ii) a non-monotonic increased capacity behavior where the hiding capacity initially plateaus before deteriorating. This effect may be due to the fact that each pixel of the cover image must hide information from multiple secret images.
We argue that this approach is suboptimal due to its inability to fully exploit the redundancy present in the cover image and secret images. This oversight fundamentally limits further improvement in hiding capacity of conventional steganography approaches. 
Therefore, the above phenomenon underscores the critical need for new hiding strategies that maintain effective capacity-distortion trade-off.
Note that, the proposed approach SMILENet, in contrast, demonstrates improved trade-off characteristics under the same multi-image hiding paradigm.

In this paper, we propose a novel Synergistic Mosaic INvertible  Image Hiding Network (SMILENet) for \textit{extra-large capacity} image steganography. Our motivation is to leverage the inherent information redundancy present in secret and cover images by synergistically combining CNNs’ non-invertibility and INNs’ invertibility to \textit{select, transform} and \textit{hide} the information of secret images effectively. In the hiding process, a Secret Information Selection (SIS) module based on the non-reversible CNNs selectively passes essential secret information to be hidden in the cover image, thereby making better use of the cover image's hiding redundancy.
Next, an Invertible Cover-Driven Mosaic (ICDM) module first reversibly transforms the selected secret information into secret representations with a reduced spatial resolution under the guidance of the cover image. It then spatially splices all secret representations to a Mosaic Secret Representation (MSR) that matches the spatial dimensions of the cover image. This design minimizes information interference among secret images while enhancing both hiding capacity and robustness. Following this, an Invertible Mosaic Secret Embedding (IMSE) module conceals the MSR within the cover image via its forward pass. 
During the recovery process, due to the reversibility of INNs, the reverse pass of the IMSE module and the ICDM module can accurately recover MSR and then reconstruct the pre-processed secret images. Finally, a Secret Detail Enhancement (SDE) module, also based on CNNs, is employed to enhance the detail information that was initially filtered out by the SIS module during the hiding process. 

The contribution of this paper is four-fold:
\begin{itemize}
    \item We propose a novel  Synergistic Mosaic InvertibLE Hiding Network (SMILENet) for extra-large capacity image steganography which fully exploits the information redundancy of cover image and secret images.

    \item The proposed SMILENet enables effective and efficient end-to-end model training with the synergistic reversible and irreversible network design and with an auxiliary loss leading to improved hiding and recovery performance.
    
    \item We propose a new  Capacity-Distortion Trade-off for comprehensively evaluating the hiding performance of image steganography methods from a perspective that considers both capacity and distortion. It also enables to compare the results of hiding various number of secret images.
    
    \item The proposed SMILENet, for the first time, achieves to hide and recover 25 secret images with a low computational complexity. Extensive experimental results show that SMILENet outperforms state-of-the-art methods in terms of hiding quality, hiding capacity and security against steganalysis methods.

\end{itemize}

\noindent \textbf{Difference from the conference paper.}
A preliminary version of this work, titled InvMIHNet, was published at ICASSP 2024~\cite{chen2024InvMIHNet}.
Beyond the conference version, we present substantial novel designs on the network architecture and training strategy that enables an effective and efficient end-to-end trainable model and leads to a further improved hiding performance and hiding capacity. Specifically, SMILENet introduces a pair of SIS module and SDE module based on CNNs to enable better utilization of the hiding redundancy of the cover image by controlling the amount of information to be embedded in the cover image and enhancing the missing details in the recovered secret image, respectively. 
Moreover, we propose a new ICDM module with a conditional invertible network architecture to transform the secret images to a mosaic secret representation which adapts to the cover image to improve hiding and recovery performance as well as robustness.
InvMIHNet is an invertible architecture with two invertible modules, but requires to train the whole network in a two-stage manner. 
In contrast, the SMILENet architecture is a synergistic model with CNNs and INNs, and enables end-to-end model training with a new hiding loss and a new auxiliary loss. 
Furthermore, we propose Capacity-Distortion Trade-off as a new metric for
evaluating image steganography algorithms from a perspective of considering both hiding capacity and distortion. 
Experiments on three datasets with different image resolutions show that SMILENet improves the hiding and recovery performances on all settings and the anti-steganalysis ability over InvMIHNet by a large margin.

The rest of paper is organized as follows: Section \ref{sec:relataed works} reviews the related work on image steganography. Section \ref{sec:proposed_method} introduces the proposed SMILENet method in details and Section \ref{sec:experiments} presents comprehensive experimental results. Finally, we draw conclusions in Section \ref{sec:conclusion}.

\section{Related works}
\label{sec:relataed works}
\begin{figure}[t]
    \centering
    \includegraphics[width=0.95\linewidth]{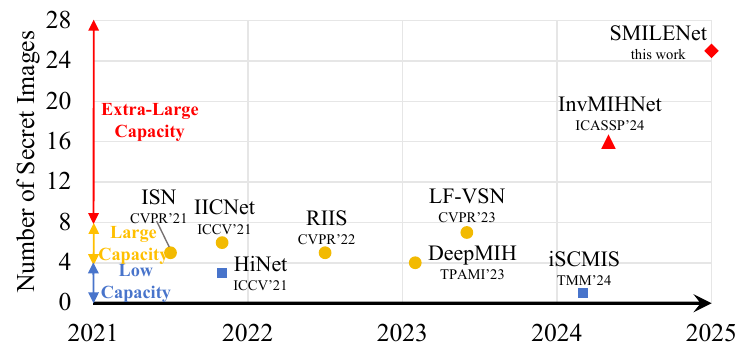}    
    \caption{The hiding capability of the representative image steganography methods in terms of the number of secret images. 
    }
    \label{fig:Comparison-Capacity}
\end{figure}

This section briefly reviews related image steganography methods based on Convolutional Neural Networks (CNNs) and Invertible Neural Networks (INNs). Fig.~\ref{fig:Comparison-Capacity} illustrates the hiding capacity of typical image steganography methods in terms of number of secret images since 2021. Here, we define the capability of hiding and recovery secret images into three categories: low capacity ($1 \sim 4$ secret images), large capacity ($5 \sim 8$ secret images) and extra-large capacity (over 9 secret images).

\subsection{CNN-based Image Steganography Methods}
The CNN-based image steganography methods have been proposed and achieved improved hiding capacity compared to traditional image steganography methods. They typically can achieve low capacity hiding  for 1 to 3 secret images.

Baluja~\cite{Baluja2017HidingII} first proposes to hide a full-size color secret image in a cover image using a Deep Steganography Network (DSN) with a preprocessing network, a hiding network and a recovery network. 
Then the same author~\cite{baluja2019hiding} further extends the DSN network architecture for hiding 3 secret images by introducing an exclusive hiding network and a recovery network for each secret image. 
Based on DSN~\cite{Baluja2017HidingII}, Rahim \textit{et al.}~\cite{rahim2018end} propose to add extra constraints on the learned weights for encoder and decoder networks to ensure joint training of the hiding and recovery networks. 
Zhu \textit{et al.}~\cite{Zhu2018HiDDeNHD} propose a Hiding Data with Deep Network (HiDDeN) for binary messages hiding with jointly trained encoder and decoder networks and by leveraging adversarial training to enhance the visual quality of stego images.
Weng \textit{et al.}~\cite{weng2019high} propose a video steganography deep neural network using temporal residual modeling which effectively enhances embedding capacity and concealment.
Zhang \textit{et al.}~\cite{zhang2020udh} propose a Universal Deep Hiding (UDH) method without using a preparation network, which achieves promising performance on cover-agnostic image steganography with up to 3 secret images. 
Duan \textit{et al.}~\cite{duan2020steganocnn} introduce a Steganography Convolutional Neural Network (SteganoCNN) with two distinct extraction networks, designed to embed two concatenated secret images into a cover image. 
Das \textit{et al.}~\cite{das2021multi} propose a preparation network to transform secret images to a suitable format for concatenation with cover images, along with a hiding and revealing network to facilitate both hiding and recovery of secret data.
Chen \textit{et al.}~\cite{chen2022hiding} propose a stego SinGAN method to hide secret images in deep probabilistic models with the guidance of the patch distribution of cover images modeled by a deep neural network.

The CNN-based image steganography methods have significantly improved the hiding capacity compared to traditional approaches, while it is difficult to impose the reversibility of the hiding network and the recovery network and therefore reaches a bottleneck when trying to further improve the hiding capability.

\subsection{INN-based Image Steganography Methods}

Recently, INNs have been successfully used for image steganography~\cite{Jing2021HiNetDI,Lu2021LargecapacityIS,xu2022robust,guan2022deepmih,chen2024InvMIHNet} due to their inherent reversibility.
INNs ~\cite{dinh2014nice, dinh2016density, gomez2017reversible, jacobsen2018revnet} are a class of deep networks with information preservation property and learn a reversible mapping from input manifold to the output.

Therefore, they have become a powerful tool for learning invertible non-linear mappings and have been widely used in solving bidirectional image transformation tasks~\cite{xiao2022invertible,zhang2022enhancing,ardizzone2019guided,zhao2021invertible,zhu2019residual,huang2021video,Liu_2021_CVPR, huang2021linn,huang2021winnet, chen2023imperceptible}.

The INN-based image steganography methods can achieve large capacity image hiding and recovery up to 7 secret images.
Jing \textit{et al.}~\cite{Jing2021HiNetDI} first introduce a HiNet for single image steganography which significantly simplifies network structure and improves the performance of both image hiding and image recovery.
Lu \textit{et al.}~\cite{Lu2021LargecapacityIS} propose Invertible Steganography Networks (ISN) by directly channel-wise concatenating multiple secret images as a single secret input. ISN method can hide and recover 5 secret images without changing the network architecture.
Cheng \textit{et al.}~\cite{cheng2021iicnet} propose an Invertible Image Conversion Net (IICNet), which incorporates a relation module and a channel squeeze layer to enhance the nonlinearity of INNs and increase the network's flexibility. 
Guan \textit{et al.}~\cite{guan2022deepmih} propose a Deep Invertible network for Multiple Image Hiding (DeepMIH) method to sequentially hide multiple secret images into the cover image with multiple INNs and use an importance map to identify the residual embedding potentials of the cover image. 
Xu \textit{et al.}~\cite{xu2022robust} propose a Robust Invertible Image Steganography (RIIS) method which considers distortions including JPEG compression, random noise and Poisson noise.
Mou \textit{et al.}~\cite{Mou2024LFVSN} propose a Large-capacity and Flexible Video Steganography Network (LF-VSN) for video steganography which can hide and recover 7 secret videos on one cover video.
Building on Guan's work~\cite{guan2022deepmih}, Li \textit{et al.}~\cite{li2023iscmis} introduce an iSCMIS method to sequentially hide and recover secret images with multiple Spatial-Channel attention based Deep Invertible Networks and demonstrate the capability to hide and recover 3 secret images.
Recently, Chen \textit{et al.}~\cite{chen2024InvMIHNet} propose an Invertible Mosaic Image Hiding Network (InvMIHNet) for very large capacity image steganography with a two-stage training strategy. It exploits the spatial redundancy of the cover image as well as the secret images and improves the hiding capacity to up to 16 secret images. 

While INN-based image steganography methods have further improved the hiding capacity, they usually overlooked the importance of exploiting the information redundancy of both cover and secret images.

\section{Proposed Method}
\label{sec:proposed_method}

In this section, we first revisit the definition of hiding capacity from a capacity-distortion perspective, then give a comprehensive overview of the proposed Synergistic Mosaic InvertibLE Hiding Network (SMILENet), followed by a detailed presentation of the network architecture design and training strategy. 
For clarity, we note some frequently used notations in Table \ref{tab:notation}.

\subsection{Revisiting Hiding Capacity from a Capacity-Distortion Perspective}

In current image steganography literature, the performance evaluation of multi-image hiding algorithms is typically assessed through two comparisons: (i) between the cover and stego image pair, and (ii) between the input and recovered secret image pairs. Given the same number of secret images $N$, these evaluations employ standard image quality metrics including Peak Signal-to-Noise Ratio (PSNR), Structural Similarity Index (SSIM)~\cite{wang2004image}, and Root Mean Square Error (RMSE).

However, this approach primarily evaluates the ``hiding capacity" based on the distortion introduced during the hiding and recovery processes. This leaves a significant gap in establishing a more rigorous and scientifically grounded measure of ``capacity" that accounts for factors beyond distortion, such as information-theoretic limits.

From the perspective of image compression, learning-based image steganography approaches can be viewed as a form of lossy compression for secret images, where the secret images are first compressed and then embedded within a cover image, and subsequently recovered from the stego image. 
This compression analogy naturally suggests a connection to Rate-Distortion (RD) theory~\cite{shannon1959coding,cover1999elements}, a framework used in lossy image compression to analyze the trade-off between the bitrate required for data representation and the distortion incurred during the reconstruction. 

\begin{table}[t]
    \caption{{The notations used in this paper. }}
    \centering
    \small
    \begin{tabular}{c|L{6cm}}
    \toprule[1pt]
            \rowcolor[HTML]{EFEFEF} \textbf{Notations} & \textbf{Description} \\ 
        \midrule
        $K, W,H$ & The channels, width and height of images\\
        $N=m \times n$ & The number of secret images\\
        $C, D$ & The capacity and distortion\\
        $R$ & Number of cInv blocks in ICDM module\\ 
        $G$ & Number of Inv blocks in IMSE module\\ 
        $\{ \bm{x}_{si} \}_{i=1}^N$ & The secret images\\
        $\{ \tilde{\bm{x}}_{si} \}_{i=1}^N$ & The pre-processed secret images\\
        $\{\bm{\bar{x}}_{si} \}_{i=1}^N$ & The initially recovered secret images\\
        $\{\bm{\hat{x}}_{si} \}_{i=1}^N$ & The finally recovered secret images\\
        $\{\bm{f}_{si}\}_{i=1}^N$ & The secret representations\\
        $\{\bm{\hat{f}}_{si}\}_{i=1}^N$ & The recovered secret representations\\
        $\bm{x}_{ms}$ & The mosaic secret representation (MSR)\\
        $\bm{\hat{x}}_{ms}$ & The recovered MSR\\
        $\bm{x}_{c}$ & The cover image\\
        $\bm{\hat{x}}_{c}$ & The recovered cover image \\
        $\bm{x}_{\text{stego}}$ & The stego image\\
        $\bm{r}_{\mathcal{H}}$ & The residual information\\
        $\bm{z}_{\mathcal{H}}$ & The auxiliary variable\\
        $\mathcal{G}(\cdot)$ & The conditional feature extractor network\\
        $\mathcal{D}(\cdot)$/$\mathcal{D}^{-1}(\cdot)$ & The decomposition/recomposition block\\
        $\mathcal{H}(\cdot)$/$\mathcal{H}^{-1}(\cdot)$ & The DWT/IDWT operation\\
        $\mathcal{Q}(\cdot)$ & The quantization operator\\
        $\psi(\cdot)$/$\rho(\cdot)$/$\phi(\cdot)$ & The dense networks\\
        $\psi_c(\cdot, \cdot)$/$\rho_c(\cdot, \cdot)$ & \multirow{2}[0]{*}{The conditional dense networks}\\
        /$\phi_c(\cdot, \cdot)$\\
    \toprule[1pt]
    \end{tabular}
    
    \label{tab:notation}
\end{table}

\noindent \textbf{Capacity-Distortion Trade-off}:
Inspired by the Rate-Distortion theory~\cite{shannon1959coding}, in this work we advocate for the adoption of Capacity-Distortion Trade-off as a new metric for evaluating image steganography algorithms. This trade-off characterizes the maximal achievable capacity $C$ for any given distortion $D$.
The (information) Capacity-Distortion function $C(D)$ is expressed as:
\begin{equation}
    \begin{aligned}
        &C(D) = \underset{P_{\hat{\bm{x}}_{si} | \bm{x}_{si}}}{\max} \sum_{i=1}^{N} I(\bm{x}_{si}, \hat{\bm{x}}_{si}),\\
        & \text{s.t. } \mathbb{E}[\Delta(\bm{x}_c, \hat{\bm{x}}_c)] \leq D,
    \end{aligned}
    \label{eq:CD_fun}
\end{equation}
where $I(\bm{x}_{si}, \hat{\bm{x}}_{si})$ denotes the normalized mutual information~\cite{cover1999elements} between the $i$-th original and reconstructed secret image, $N$ denotes the number of secret images hiding in the cover image, $\Delta: \mathcal{X} \times \hat{\mathcal{X}} \rightarrow \mathcal{R}^{+}$ is any full reference distortion measure such that $\Delta(\bm{x}_c, \hat{\bm{x}}_c)=0$ iff $\bm{x}_c=\hat{\bm{x}}_c$, and $D$ is used to bound the expected distortion between the original and reconstructed cover image.

The Capacity-Distortion framework establishes an inherent information-theoretic capacity (maximum payload size) and visual distortion (quantified through metrics like RMSE/SSIM). 
Under this paradigm, a method that achieves higher hiding capacity while maintaining lower distortion is considered superior. 
This framework offers a comprehensive and unified approach for evaluating the performance of image steganography methods across varying number of secret images.
By capturing the trade-off between capacity and distortion, it offers a more nuanced and theoretically grounded measure of a method's overall performance in image steganography.
From Fig.~\ref{fig:capacity_rmse_SMILENet} we can see that, under the Capacity-Distortion paradigm, the proposed approach SMILENet, demonstrates improved trade-off characteristics under the same multi-image hiding paradigm when comparing with ISN~\cite{Lu2021LargecapacityIS} and InvMIHNet~\cite{chen2024InvMIHNet}.

\begin{figure*}[t]
    \centering
    \includegraphics[width=0.85\linewidth]{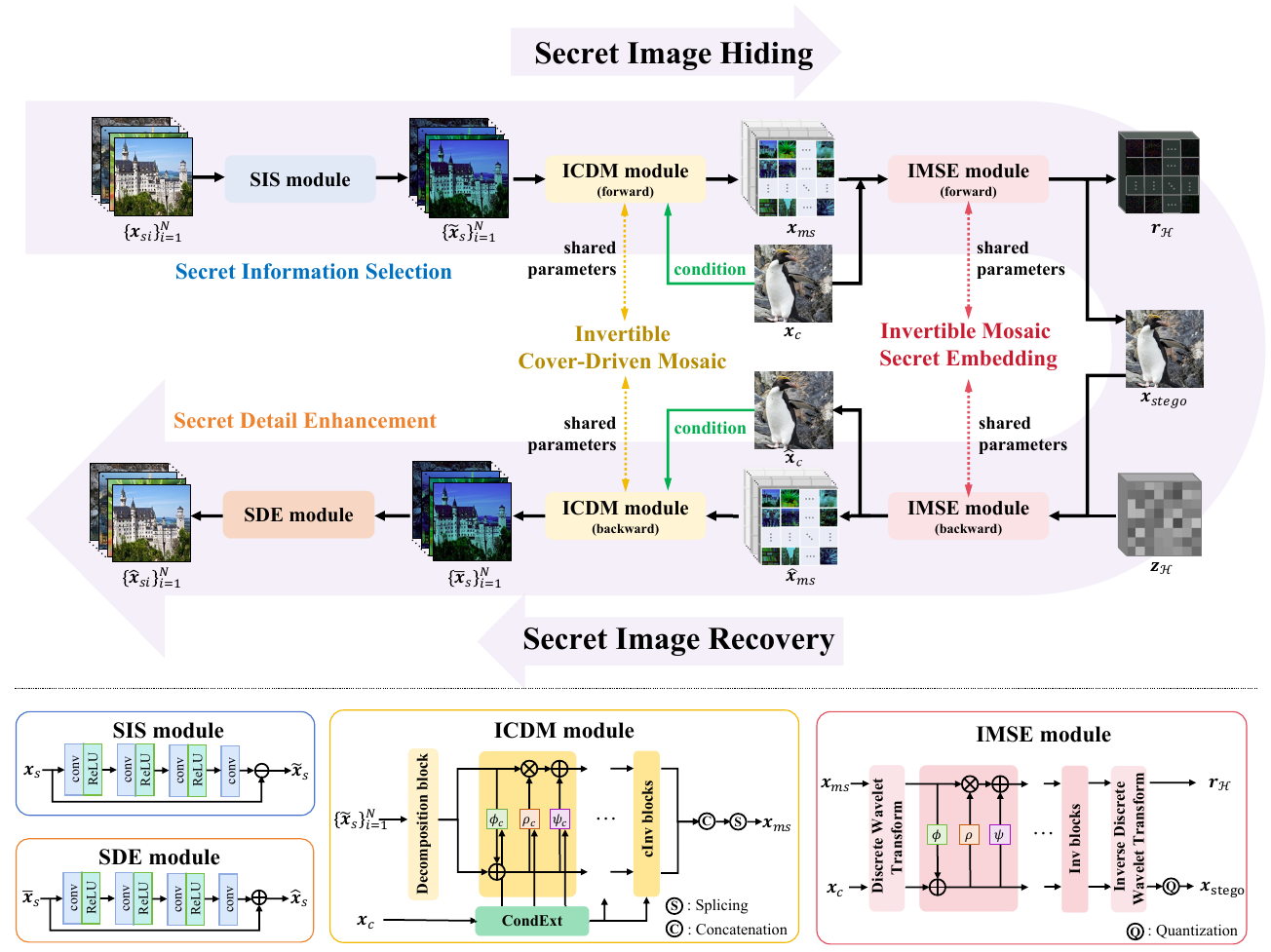}
    \caption{Framework overview of the proposed SMILENet. 
    During the hiding process, the secret images $\{ \bm{x}_{si} \}_{i=1}^N$ are first processed by an SIS module to select essential information $\{ \bm{\tilde{x}}_{si} \}_{i=1}^N$ to be concealed. Next, an ICDM module, guided by the cover image, reversibly transforms $\{ \bm{\tilde{x}}_{si} \}_{i=1}^N$ to secret representations and then spatially splices them to a Mosaic Secret Representation (MSR) $\bm{x}_{ms}$. Subsequently, an IMSE module conceals the MSR into the cover image $\bm{x}_{c}$. Its outputs include a stego image $\bm{x}_{\text{stego}}$ and an image-agnostic component $\bm{r}_{\mathcal{H}}$ which is discarded thereafter. During the recovery process, the reverse pass of the IMSE module and the ICDM module recover MSR $\bm{\hat{x}}_{ms}$ and reconstruct the pre-processed secret images $\{ \bm{\bar{x}}_{si} \}_{i=1}^N$. Finally, a Secret Detail Enhancement (SDE) module refines the recovered images by enhancing their fine details, thereby generating the final secret images $\{ \bm{\hat{x}}_{si} \}_{i=1}^N$.}
    \label{fig:overview}
\end{figure*}

\subsection{Framework Overview}
\label{sec:methodology}

SMILENet is a novel architecture for extra-large capacity image steganography that synergistically leverages the non-reversibility of CNNs and the invertibility of INNs to exploit the inherent information redundancy in both secret and cover images. 
SMILENet consists of two INNs based modules \textit{i.e.}, an Invertible Cover-Driven Mosaic (ICDM) module and an Invertible Mosaic Secret Embedding (IMSE) module, together with two CNNs based modules \textit{i.e.}, a Secret Information Selection (SIS) module and a Secret Detail Enhancement (SDE) module.
Specifically, the ICDM module, which is a conditional INN guided by the cover image, adaptively transforms multiple secret images into a Mosaic Secret Representation (MSR) during the forward pass and recovers them during the reverse pass, thereby minimizing interference among secret images and enhancing hiding capacity and robustness. The IMSE module faithfully conceals the MSR into the cover image during the forward pass and recovers the MSR from the stego image during the backward pass. The SIS module selects essential information in secret images to fully exploit the hiding redundancy of the cover image, and the SDE module enriches the details in the recovered secret images.

The framework of the proposed SMILENet is given in Fig.~\ref{fig:overview}, where we take the case of hiding $N=m \times n$ ($m, n \in \mathbb{Z}$) images for illustration. 
We assume that the cover image $\bm{x}_c$ and $N$ secret images $\{ \bm{x}_{si} \}_{i=1}^N$ are of the same size $K \times W \times H$.
Each secret image $\bm{x}_{si}$ is first pre-processed by the SIS module where its essential information is selected to be hidden in the cover image.
The subsequent hiding process is modeled as the forward process of the ICDM module and the IMSE module.
The ICDM module, guided by the cover image, reversibly transforms secret representations $\{\bm{x}_{si} \}_{i=1}^N$ which are then spliced into $m$ rows and $n$ columns to form a Mosaic Secret Representation (MSR).
Subsequently, the IMSE module's forward pass hides MSR into the cover image. The recovery process is modeled as the reverse pass  through the IMSE module and the ICDM module. This process sequentially and reversibly recovers the MSR $\hat{\bm{x}}_{ms}$ and the cover image $\hat{\bm{x}}_c$, and then transforms the MSR back into the initially recovered secret images, guided by the recovered cover image.
Finally, the SDE module is employed to enhanced the essential details of the recovered secret images $\{\bm{\hat{x}}_{si} \}_{i=1}^N$.
The overall SMILENet can be trained in an end-to-end manner, allowing for joint optimization of all modules. 

In the following, we introduce SMILENet in detail by first describing the hiding process and then the recovery process.

\subsection{Secret Image Hiding}

\Tianruinew{The secret image hiding pathway consists of three core components: the Secret Information Selection (SIS) module, the Invertible Cover-Driven Mosaic (ICDM) module, and the Invertible Mosaic Secret Embedding (IMSE) module. These components work synergistically to select, transform, and hide multiple secret images into a cover image while minimizing information interference and maximizing redundancy utilization.  }

\subsubsection{Secret Information Selection (SIS) Module}

The SIS module selectively identifies and passes only essential secret information to the subsequent ICDM module and IMSE module, ensuring efficient utilization of the hiding redundancy of the cover image.
\Tianruinew{As shown in Fig.~\ref{fig:overview},} the SIS module comprises $L$ convolutional layers integrated with a residual connection which is designed to detect and filter out redundant information that can be safely omitted without compromising recovery performance.
The SIS module not only optimizes the use of available capacity within the cover image but also minimizes potential distortions, leading to higher fidelity in the stego image.

\subsubsection{\Tianruinew{Invertible Cover-Driven Mosaic (ICDM) Module}}
\Tianruinew{
The proposed ICDM module, structured as a conditional Invertible Neural Network (cINN), leverages the cover image as a dynamic condition to adaptively transform preprocessed secret images into a compact Mosaic Secret Representation (MSR). }
This design allows the following module to adapt the secret representation to the cover image, ensuring that the hiding performance is independent of the hiding order, thereby enhances both hiding and recovery performance, as well as overall robustness.

\noindent \textbf{The Forward Pass of ICDM:} 
The forward pass of {ICDM} consists of a decomposition block $\mathcal{D}(\cdot)$ and a sequence of $R$ conditional Invertible  (cInv) blocks. To flexibly rescale the secret images by any factor of $m \times n$, the decomposition block $\mathcal{D}(\cdot)$ incorporates an invertible convolution layer that transforms a pre-processed secret image $\tilde{\bm{x}}_{si} \in \mathbb{R}^{K \times W \times H}$, into a decomposed representation $\mathcal{D}(\tilde{\bm{x}}_{si}) \in \mathbb{R}^{mnK \times W/m \times H/n}$. The invertible convolution layer is parameterized by the Cayley transform~\cite{absil2009optimization, huang2021winnet} and employs $mn$ filters, each with a kernel size and stride equal to $m \times n$. 
The invertible convolutional kernel is set to a learned orthogonal matrix $\bm{K}$:
\begin{equation}
    \bm{K} = \left(\mathbf{I} - (\bm{\Theta} - \bm{\Theta}^T)\right) \left(\mathbf{I} + (\bm{\Theta} - \bm{\Theta}^T)\right)^{-1},
    \label{eq:Cayley}
\end{equation}
where $\mathbf{I}$ is an identity matrix, and $\bm{\Theta} \in \mathbb{R}^{N \times N}$ denotes learnable parameters.

The decomposed representation $\mathcal{D}({\tilde{\bm{x}}}_{si})$ is split into a top part $\bm{f}_{ti}^{0}  \in \mathbb{R}^{(mn-1)K \times W/m \times H/n}$ 
and a bottom part $\bm{f}_{bi}^{0} \in \mathbb{R}^{K \times W/m \times H/n}$. The top and bottom features are alternatively updated with $R$ cInv blocks which are guided by the conditional feature $\mathcal{G}(\bm{x}_c)$ extracted from the cover image. The forward process of the $r$-th cInv block can be expressed as:
\begin{equation}
    \begin{cases}
    \bm{f}_{bi}^{r} &= \bm{f}_{bi}^{r-1}+{\phi_c}({\bm{f}_{ti}^{r-1};\mathcal{G}(\bm{x}_{c})}),\\
    \bm{f}_{ti}^{r} &= \bm{f}_{ti}^{r-1} \odot \exp \left(\rho_c \left (\bm{f}_{bi}^{r};\mathcal{G}(\bm{x}_{c})\right) \right) + {\psi_c}(\bm{f}_{bi}^{r};\mathcal{G}(\bm{x}_{c})),
    \end{cases}
    \label{eq:CInv_forward}
\end{equation}
where the subscript $b$ and $t$ denote the bottom and top contents, $\odot$ denotes element-wise multiplication, $\phi_c(\cdot, \cdot), \rho_c(\cdot, \cdot)$ and $\psi_c(\cdot, \cdot)$ represent dense networks~\cite{huang2017densely} with a conditional input, and $\mathcal{G}(\cdot)$ represents conditional feature extractor. 

\begin{figure}
    \centering
    \includegraphics[width=1.0\linewidth]{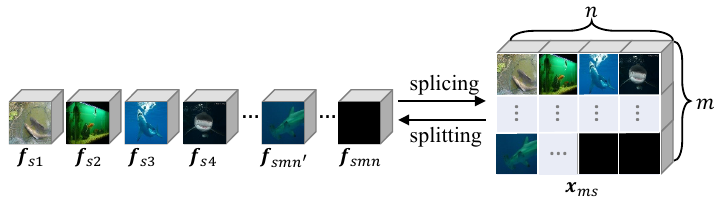}
    \caption{The process of splicing and splitting  $N=m \times n$ secret images.}
    \label{fig:splicing_splitting}
    \vspace{-3pt}
\end{figure}

At the end of the ICDM module, the updated top and bottom features $\{\bm{f}_{bi}^{R}, \bm{f}_{ti}^{R}\}$ of the $i$-th secret image are concatenated to form the secret representation $\bm{f}_{si}$. 
The secret image splicing operation then arranges all the secret representations $\{ \bm{f}_{si} \}_{i=1}^N$ to construct the Mosaic Secret Representation (MSR), as illustrated in Fig.~\ref{fig:splicing_splitting}. 

\begin{itemize}
    \item When $N$ is a composite number, it is decomposed as $N=m \times n$ with $m,n \in \mathbb{Z}$ such that $|m-n|$ is minimized. The $N$ secret representations $\{ \bm{f}_{si} \}_{i=1}^N$ are then arranged in a grid with $m$ rows and $n$ columns to form the MSR $\bm{x}_{ms}$. 

    \item When $N$ is a prime number, we identify the nearest larger composite number $N'$. 
    We follow the same decomposition principle $N'=m \times n$, and arrange $\{ \bm{f}_{si} \}_{i=1}^N$ alongside  $(N'-N)$ all-zero images into a grid with $m$ rows and $n$ columns. 
\end{itemize}

This approach ensures that the MSR retains a uniform structure while providing high flexibility in handling varying numbers of images.

\subsubsection{\Tianruinew{Invertible Mosaic Secret Embedding (IMSE)} Module}
\Tianruinew{The proposed IMSE module, following the ICDM module, embeds the MSR into the cover image via wavelet-based invertible operations and a quantization operator. By leveraging the inherent invertibility of INNs, the IMSE module achieves high-fidelity embedding, ensuring reliable secret recovery with minimal distortion.}

\noindent \textbf{The Forward Pass of IMSE:} 
In the forward pass of the IMSE module, the DWT block first transforms both $\bm{x}_{ms}$ and the cover image $\bm{x}_{c}$ into their corresponding wavelet representations, denoted as  $\bm{f}_{ms}^{0}$ and $\bm{f}_{c}^{0}$, respectively. This transformation enables a deeper feature interactions in the subsequent $G$ Inv blocks which exchanges information between the features of the MSR and the cover image.
The forward process of the $g$-th Inv blocks can be expressed as:
\begin{equation}
    \begin{cases}
    \bm{f}_{ms}^{g} &= \bm{f}_{ms}^{g-1}+{\phi}({\bm{f}_{c}^{g-1}}),\\
    \bm{f}_{c}^{g} &= \bm{f}_{c}^{g-1} \odot \exp \left(\rho \left (\bm{f}_{ms}^{g}\right) \right) + {\psi}(\bm{f}_{ms}^{g}),
    \end{cases}
    \label{eq:Inv_forward}
\end{equation}
where $\bm{f}_{ms}^{0} = \mathcal{H}(\bm{x}_{ms}) $ and $ \bm{f}_{c}^{0} = \mathcal{H}(\bm{x}_{c})$ with $\mathcal{H}(\cdot)$ denoting the DWT operation, and Haar Wavelet transform is applied. Moreover, $\phi(\cdot), \rho(\cdot)$ and $\psi(\cdot)$ represent dense networks~\cite{huang2017densely}.

The stego image is then obtained by sequentially passing $\bm{f}_{c}^{G}$ through the IDWT block to transform it back to image domain, followed by the quantization operator to simulate real-world conditions:
\begin{equation}
    \bm{x}_{\text{stego}} = Q(\mathcal{H}^{-1}(\bm{f}_{c}^{G})),
\end{equation}
where $\mathcal{H}^{-1}(\cdot)$ denotes the the IDWT block, and $Q(\cdot)$ denotes the quantization operator.
The quantization operator clamps the pixels values by rounding pixel values into integers within the range of $[0, 255]$ to simulate 8-bit images. Although quantization is inherently non-differentiable, we employ  the Straight-Through Estimator~\cite{bengio2013estimating}.
In this approach, uniform noise is added during training to approximate gradients, while integer rounding is applied during testing.
The updated mosaic feature $\bm{r}_{\mathcal{H}}  = \mathcal{H}^{-1}(\bm{f}_{ms}^{G})$ is treated as image-agnostic variable and is discarded.

\subsection{Secret Image Recovery}

\Tianruinew{The secret image recovery pathway also comprises three key stages, \textit{i.e.,} the reverse precess of the IMSE and ICDM module, as well as the Secret Detail Enhancement (SDE) module. 
}

\subsubsection{Invertible Cover-Driven Mosaic (ICDM) Module}
The reverse pass of IMSE first reconstructs the mosaic secret representation from the stego image through inverse wavelet transformations and invertible block operations.

\noindent \textbf{The Reverse Pass of IMSE:} 
The reverse pass of the ICDM module consists of the reverse process of the $R$ cInv blocks followed by a recomposition block $\mathcal{D}^{-1}(\cdot)$. The recovered cover images and recovered MSR $\bm{\hat{x}}_{ms}$ can be obtained from the stego image $\bm{x}_{\text{stego}}$.
Specifically, the stego image $\bm{x}_{\text{stego}}$ and a sampled random variable $\bm{z}_{\mathcal{H}}$ are input to the reverse pass of the $G$ Inv blocks to restore the cover image and the recovered MSR. 
By construction, this process is fully reversible and the reverse process of the $g$-th Inv block can be expressed as:
\begin{equation}
    \begin{cases}
        \bm{\hat{f}}_{c}^{g-1} &=  (\bm{\hat{f}}_{c}^{g}-{\psi} (\bm{\hat{f}}_{ms}^{g} ) ) \odot \exp (-\rho (\bm{\hat{f}}_{ms}^{g} ) ),\\
        \bm{\hat{f}}_{ms}^{g-1} &= \bm{\hat{f}}_{ms}^{g}-{\phi}({\bm{\hat{f}}_{c}^{g-1}}),
    \end{cases}
    \label{eq:Inv_backward}
\end{equation}
where ${\hat{f}}_{ms}^{G} = \mathcal{H}(\bm{z}_{\mathcal{H}})$ and ${\hat{f}}_{c}^{G} = \mathcal{H}(\bm{x}_{\text{stego}})$.

The IDWT block then transforms $\bm{\hat{f}}_{ms}^{0}$,  $\bm{\hat{f}}_{c}^{0}$ to the recovered MSR and the recovered cover image: $\bm{\hat{x}}_{ms} = \mathcal{H}^{-1}(\bm{\hat{f}}_{ms}^{0}) $ and $ \bm{\hat{x}}_{c} = \mathcal{H}^{-1}(\bm{\hat{f}}_{c}^{0})$.

\subsubsection{Invertible Cover-Driven Mosaic (ICDM) Module}
The reverse pass of ICDM module then decomposes the recovered MSR into individual secret images guided by the restored cover image.

\noindent \textbf{The Reverse Pass of ICDM:} 
The recovered MSR $\bm{\hat{x}}_{ms}$, is first split into 
$mn$ recovered secret representations $\{ \bm{\hat{f}}_{si} \}_{i=1}^N$, each with spatial dimensions  $W/m \times H/n$. Next, each secret representation is divided into
bottom part $\bm{\hat{f}}_{bi}^R  \in \mathbb{R}^{K \times W/m \times H/n}$ and top part $\bm{\hat{f}}_{ti}^R  \in \mathbb{R}^{(mn-1)K \times W/m \times H/n}$. These components are then updated using the reverse process of the $R$ cInv blocks, with the guidance provided by the recovered cover image $\bm{\hat{x}}_{c}$ using the conditional feature extractor $\mathcal{G}(\cdot)$:
\begin{equation}
    \begin{cases}
        \bm{\hat{f}}_{ti}^{r-1} &=  (\bm{\hat{f}}_{ti}^{r}-{\psi_c} (\bm{\hat{f}}_{bi}^{r} ; \mathcal{G}(\bm{\hat{x}}_{c})) ) \odot \exp (-\rho_c (\bm{\hat{f}}_{bi}^{r} ; \mathcal{G}(\bm{\hat{x}}_{c})) ),\\
        \bm{\hat{f}}_{bi}^{r-1} &= \bm{\hat{f}}_{bi}^{r}-{\phi_c} ({\bm{\hat{x} }_{ti}^{r-1}}; \mathcal{G}(\bm{\hat{x}}_{c})).
    \end{cases}
    \label{eq:CInv_backward}
\end{equation}

The recomposition block then obtains the recovered pre-processed secret image via $\bm{\bar{x}}_{si} = \mathcal{D}^{-1}(\bm{\hat{f}}_{ti}^{0} \copyright \bm{\hat{f}}_{bi}^{0})  \in \mathbb{R}^{K \times W \times H}$ with $\copyright$ being the concatenation operator. It is parameterized by the same orthogonal matrix as the decomposition block, thus leading by construction to invertibility.

\subsubsection{Secret Detail Enhancement (SDE) Module}
The SDE module is proposed to compensate for the loss of details in SIS module to ensure the recovery of high-quality secret images after the reverse pass through the ICDM and IMSE module.
\Tianruinew{As given in Fig.~\ref{fig:overview},} the SDE module is designed based on CNNs, and has a symmetric design mirroring that of the SIS module. \Tianruinew{Section \ref{tab:ablation_key_module} validates the contribution of SDE module to performance improvement through quantitative analysis.}

\subsection{Training Strategy and Loss Functions}
\label{sec:loss}

The overall loss function consists of three main groups: the secret loss $\mathcal{L}_{sec}$ guides the learning of secret images, the hiding loss $\mathcal{L}_{hide}$ improves the quality of the stego image, the auxiliary loss $\mathcal{L}_{aux}$ ensures that the training process of IMSE module can be optimized effectively:
\begin{equation}
    \mathcal{L}_{{total}}= \mathcal{L}_{sec} + \mathcal{L}_{hide} + \mathcal{L}_{aux}.
    \label{eq:totalloss}
\end{equation}

In specific, the secret loss $\mathcal{L}_{sec}$ is used to guarantee the quality of the recovered secret images $\{ \bm{\hat{x}}_{si} \}_{i=1}^{N}$ resemble to that of the original secret images $\{ \bm{x}_{si}\}_{i=1}^{N}$.
It is expressed as:
\begin{equation}
    \begin{aligned}
        \mathcal{L}_{sec} = \sum_{i=1}^{N} \left\|\bm{x}_{si}-\bm{\hat{x}}_{si} \right\|_1.
    \end{aligned}
    \label{equ:secret_loss}
\end{equation}

The hiding loss $\mathcal{L}_{hide}$ is designed to ensure that the stego images $\bm{x}_{\text{stego}}$ are indistinguishable from the cover images $\bm{x}_{c}$, especially on the low-frequency subband:
\begin{equation}
    \begin{aligned}
        \mathcal{L}_{hide} 
        =\lambda_{h} \|\bm{x}_{c} - \bm{x}_{\text{stego}}  \|^{2}_{2} +  \lambda_{hl} \| \bm{x}^{ll}_{c}-\bm{x}^{ll}_{\text{stego}} \| ^{2}_{2},
    \end{aligned}
\end{equation}
where $\lambda_{h}$ and $\lambda_{hl}$ denote the regularization parameters and superscript $ll$ denotes the low-frequency component of the wavelet transform.

The auxiliary loss $\mathcal{L}_{aux}$ includes two key constraints to optimize the performance of hiding and recovery. Specifically, $\mathcal{L}_{ms}$ aims to ensure that the backward pass of IMSE module can recover high quality MSR $\bm{\hat{x}}_{ms}$. 
Meanwhile, $\mathcal{L}_{rc}$ restricts the recovered cover images $\bm{\hat{x}}_{c}$ by the IMSE module to remain  similar to the original cover images $\bm{x}_{c}$, therefore ensuring the recover quality of the ICDM module. It is expressed as:
\begin{equation}
    \begin{aligned}
    \mathcal{L}_{aux} &= \mathcal{L}_{ms}+\mathcal{L}_{rc},\\
    &=\lambda_{ms} \left \| \bm{x}_{ms}-\bm{\hat{x}}_{ms} \right \|^{2}_{2} + {\lambda_{rc} \|\bm{x}_{c}-\bm{\hat{x}}_{c}\|^{2}_{2}},
    \end{aligned}
\end{equation}
where $\lambda_{ms}$ and $\lambda_{rc}$ denote the regularization parameters for balancing hiding and recovery of IMSE module.

\begin{table}[t]
    \centering
    \renewcommand\arraystretch{1.27}
    \caption{The datasets used for training and testing.}
    \begin{tabular}{c|c|c|c}
          \toprule[1pt]
          \rowcolor[HTML]{EFEFEF}  \textbf{Dataset} & \textbf{Resolution} & \textbf{Training} & \textbf{Testing} \\ 
          \midrule
          DIV2K~\cite{agustsson2017ntire} & $1024 \times 1024$ & 800 & 100\\
          Paris StreetView~\cite{doersch2015makes} & $512 \times 512$ & \XSolidBrush & 6300\\
          ImageNet~\cite{russakovsky2015imagenet} & $256 \times 256$ & \XSolidBrush & 1000\\
          \toprule[1pt]
    \end{tabular}
    
    \label{tab:dataset}
\end{table}

\begin{table*}[t]
\renewcommand\arraystretch{1.27}
  \centering
  \newcolumntype{C}[1]{>{\centering\let\newline\\\arraybackslash\hspace{0pt}}m{#1}}
  \caption{
  The hiding and recovery results of different methods on extra-large capacity setting with $N=9, 16, 25$ images evaluated on DIV2K, Paris StreetView and ImageNet datasets.  (The best results are in bold.)}
  \scalebox{0.92}{
    \begin{tabular}{l|cccc|cccc|cccc}
    \hline
    \toprule[1pt]
    \rowcolor[HTML]{EFEFEF} 
    \cellcolor[HTML]{EFEFEF}   &  \multicolumn{12}{c}{\textbf{Cover / Stego}} \\
    \cline{2-13} 
    \rowcolor[HTML]{EFEFEF} &  \multicolumn{4}{c|}{\textbf{DIV2K} ($1024 \times 1024$)} & \multicolumn{4}{c|}{\textbf{Paris StreetView} ($512 \times 512$)} & \multicolumn{4}{c}{\textbf{ImageNet} ($256 \times 256$)} \\
    \cline{2-13} 
    \rowcolor[HTML]{EFEFEF} \multirow{-3}{*}{{\textbf{Methods}}} &\textbf{PSNR} $\uparrow$ & \textbf{SSIM} $\uparrow$ & \textbf{RMSE} $\downarrow$ &\multicolumn{1}{c|}{\textbf{MAE} $\downarrow$}  &\textbf{PSNR} $\uparrow$ & \textbf{SSIM} $\uparrow$ & \textbf{RMSE} $\downarrow$ &\multicolumn{1}{c|}{\textbf{MAE} $\downarrow$}  &\textbf{PSNR} $\uparrow$ & \textbf{SSIM} $\uparrow$ & \textbf{RMSE} $\downarrow$ &\multicolumn{1}{c}{\textbf{MAE} $\downarrow$} \\
    \hline

    ISN$_{(9)}$~\cite{Lu2021LargecapacityIS}&32.14&0.916&6.53&4.57 &32.75&0.919&6.08&4.50&31.69&0.925&7.10&5.13\\
    InvMIHNet$_{(9)}$~\cite{chen2024InvMIHNet} &33.56&0.874&5.36&4.07 &33.63&0.875&5.36&4.03 &32.61&0.874&6.05&4.55\\
    \textbf{SMILENet$_{(9)}$} &\textbf{40.91}&\textbf{0.972}&\textbf{2.30}&\textbf{1.66} &\textbf{41.62}&\textbf{0.977}&\textbf{2.15}&\textbf{1.51} &\textbf{39.38}&\textbf{0.970}&\textbf{2.84}&\textbf{2.04}\\

    \rowcolor[HTML]{EFEFEF}
    ISN$_{(16)}$~\cite{Lu2021LargecapacityIS} &30.31&0.851&7.86&6.02 &29.21&0.832&10.14&7.19 &27.85&0.851&11.38&7.98\\
    \rowcolor[HTML]{EFEFEF}
    InvMIHNet$_{(16)}$~\cite{chen2024InvMIHNet} &34.57&0.919&4.78&3.49 &35.61&0.906&4.26&3.18 &33.80&0.902&5.25&3.92\\
    \rowcolor[HTML]{EFEFEF}
    \textbf{SMILENet$_{(16)}$} &\textbf{40.42}&\textbf{0.980}&\textbf{2.45}&\textbf{1.70} &\textbf{42.44}&\textbf{0.981}&\textbf{1.95}&\textbf{1.34} &\textbf{39.95}&\textbf{0.976}&\textbf{2.69}&\textbf{1.92}\\

    ISN$_{(25)}$~\cite{Lu2021LargecapacityIS} &24.23&0.856&19.12&11.45 &26.60&0.902&13.81&10.03 &25.38&0.892&15.87&11.25\\
    InvMIHNet$_{(25)}$~\cite{chen2024InvMIHNet} &-&-&-&- &-&-&-&- &-&-&-&-\\
    \textbf{SMILENet$_{(25)}$} &\textbf{41.26}&\textbf{0.966}&\textbf{2.21}&\textbf{1.63} &\textbf{41.33}&\textbf{0.977}&\textbf{2.22}&\textbf{1.58} &\textbf{38.92}&\textbf{0.964}&\textbf{3.01}&\textbf{2.18} \\

    \toprule[1pt]
    \toprule[1pt]
   \rowcolor[HTML]{EFEFEF} 
    \cellcolor[HTML]{EFEFEF}   &  \multicolumn{12}{c}{\textbf{Secret / Recovery}} \\
    \cline{2-13} 
    \rowcolor[HTML]{EFEFEF} &  \multicolumn{4}{c|}{\textbf{DIV2K} ($1024 \times 1024$)} & \multicolumn{4}{c|}{\textbf{Paris StreetView} ($512 \times 512$)} & \multicolumn{4}{c}{\textbf{ImageNet} ($256 \times 256$)} \\
    \cline{2-13} 
    \rowcolor[HTML]{EFEFEF} \multirow{-3}{*}{{\textbf{Methods}}} &\textbf{PSNR} $\uparrow$ & \textbf{SSIM} $\uparrow$ & \textbf{RMSE} $\downarrow$ &\multicolumn{1}{c|}{\textbf{MAE} $\downarrow$}  &\textbf{PSNR} $\uparrow$ & \textbf{SSIM} $\uparrow$ & \textbf{RMSE} $\downarrow$ &\multicolumn{1}{c|}{\textbf{MAE} $\downarrow$}  &\textbf{PSNR} $\uparrow$ & \textbf{SSIM} $\uparrow$ & \textbf{RMSE} $\downarrow$ &\multicolumn{1}{c}{\textbf{MAE} $\downarrow$}\\
    \hline

    ISN$_{(9)}$~\cite{Lu2021LargecapacityIS} &18.62&0.465&30.93&23.34 &19.44&0.459&28.17&21.44&18.90&0.479&30.00&23.01\\
    InvMIHNet$_{(9)}$~\cite{chen2024InvMIHNet} &30.66&0.898&8.28&5.29 &30.10&0.903&9.05&5.52 &30.86&0.891&7.96&5.17\\
    \textbf{SMILENet$_{(9)}$} &\textbf{32.21}&\textbf{0.911}&\textbf{7.02}&\textbf{4.34} &\textbf{31.62}&\textbf{0.915}&\textbf{7.79}&\textbf{4.57} &\textbf{32.08}&\textbf{0.908}&\textbf{7.01}&\textbf{4.40}\\

    \rowcolor[HTML]{EFEFEF}
    ISN$_{(16)}$~\cite{Lu2021LargecapacityIS} &18.42&0.524&32.12&23.77 &18.92 &0.536&30.68&22.72 &18.47&0.554&31.96&23.65  \\
    \rowcolor[HTML]{EFEFEF}
    InvMIHNet$_{(16)}$~\cite{chen2024InvMIHNet} &28.70&0.834&10.44&6.34 &28.41&0.841&11.13&6.52 &28.83&0.826&10.08&6.25\\
    \rowcolor[HTML]{EFEFEF}
    \textbf{SMILENet$_{(16)}$} &\textbf{30.01}&\textbf{0.861}&\textbf{9.01}&\textbf{5.44} &\textbf{29.64}&\textbf{0.868}&\textbf{9.73}&\textbf{5.62} &\textbf{30.10}&\textbf{0.862}&\textbf{8.75}&\textbf{5.39}\\

    ISN$_{(25)}$~\cite{Lu2021LargecapacityIS} &10.12&0.144&80.28&65.55 & 10.07&0.164&81.28&66.67 &10.24&0.892&15.87&11.25\\
    InvMIHNet$_{(25)}$~\cite{chen2024InvMIHNet} &-&-&-&- &-&-&-&- &-&-&-&- \\
    \textbf{SMILENet$_{(25)}$} &\textbf{28.63}&\textbf{0.823}&\textbf{10.63}&\textbf{6.35} &\textbf{28.13}&\textbf{0.822}&\textbf{11.53}&\textbf{6.66} &\textbf{28.56}&\textbf{0.822}&\textbf{10.40}&\textbf{6.41} \\

    \toprule[1pt]
    \end{tabular}%
    }
  \label{tab:very_large_capacity}%
\end{table*}%

\begin{figure*}[t]
    \centering
    \includegraphics[width=0.8\linewidth]{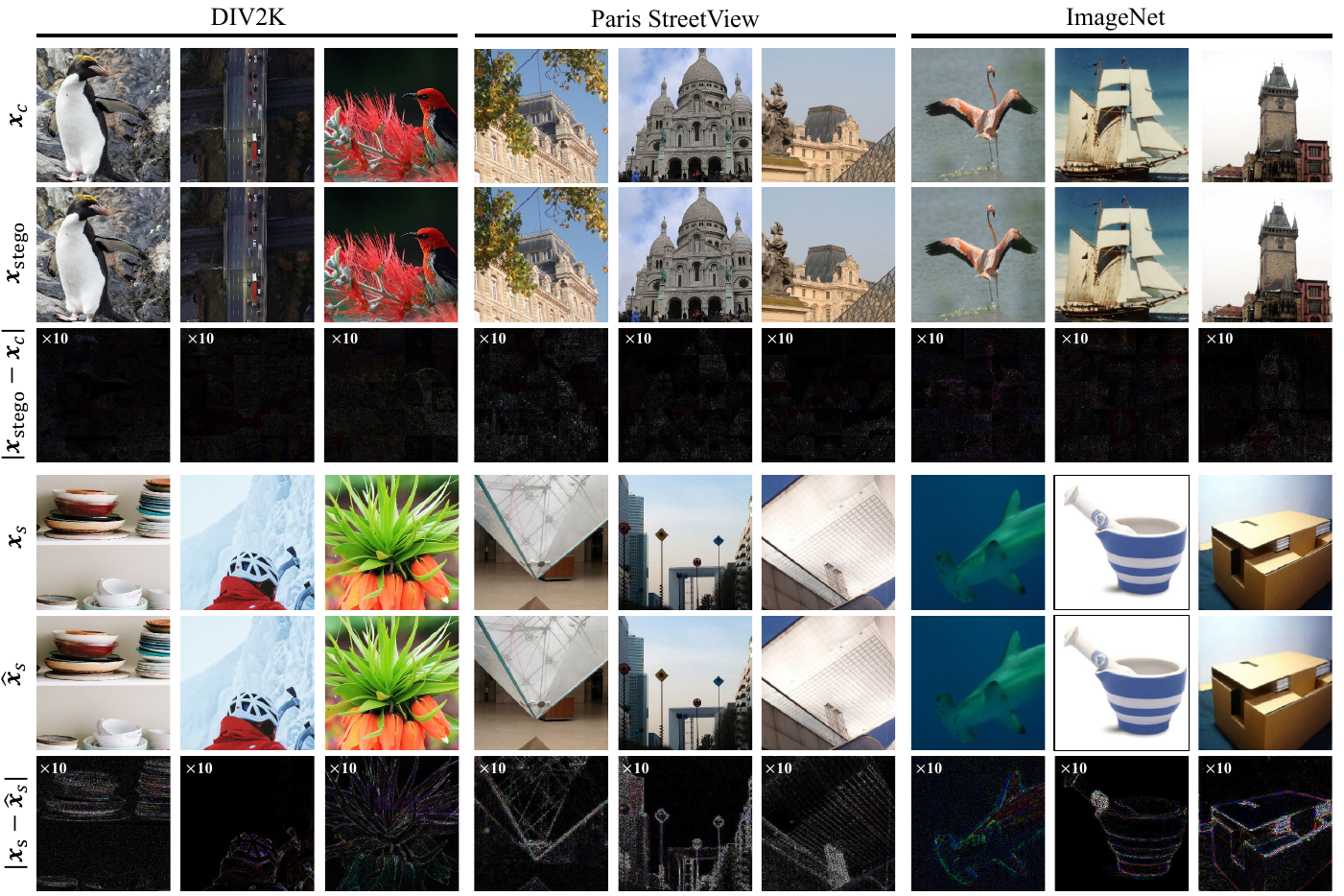}
    \caption{Visualization results of SMILENet on hiding and recovery 25 secret images evaluated on DIV2K, Paris StreetView and ImageNet datasets. 
    All residual images are enlarged 10 times for better perception.}
    \label{fig:25images}
    \vspace{-3pt}
\end{figure*}

\section{Experiments}
In this section, we present comprehensive experimental results to validate the effectiveness of the proposed SMILENet. We first introduce the experimental setup in Section \ref{sec:experimental_setup}, present evaluation results in Section \ref{sec:evaluation_results}, and the results of ablation study in Section \ref{sec:abaltion_study}.

\label{sec:experiments}
\subsection{Experimental Setup}
\label{sec:experimental_setup}
\noindent\textbf{\textbf{Dataset and settings.}} 
The datasets for training and testing are listed in Table~\ref{tab:dataset}. The DIV2K~\cite{agustsson2017ntire} training dataset contains 800 images of resolution 1024$\times$1024 and is used for training.
The testing dataset includes 100 test images from the DIV2K validation dataset~\cite{agustsson2017ntire} of resolution 1024$\times$1024, 1,000 images of resolution 256$\times$256 randomly selected from ImageNet~\cite{russakovsky2015imagenet} and
6,300 images of resolution 512$\times$ 512 from Paris StreetView dataset~\cite{doersch2015makes}.

\noindent{\textbf{Comparison methods.} }
Nine publicly available comparison methods have been included for evaluation. In specific, Weng's method~\cite{weng2019high}, Baluja's method~\cite{baluja2019hiding} and UDH~\cite{zhang2020udh} are based on Convolutional Neural Networks and can hide and recover 2 secret images. 
DeepMIH~\cite{guan2022deepmih}, 
IICNet~\cite{cheng2021iicnet}, iSCMIS~\cite{li2023iscmis}, and ISN~\cite{Lu2021LargecapacityIS} are state-of-the-art large capacity steganography methods based on Invertible Neural Networks and can hide and recover $4 \sim 5$ secret images, respectively. LF-VSN~\cite{Mou2024LFVSN} hides up to 7 videos within the cover video.
InvMIHNet~\cite{chen2024InvMIHNet} successfully hides and recovers 16 secret images.
For fair comparison, we train and test all methods on the same set of datasets.

\noindent{\textbf{Implementation details.}}
The stochastic gradient descent with Adam optimizer~\cite{kingma2014adam} is used for training with the initial learning rate $1\times10^{-4.5}$ which is halved every $100K$ iterations. The training patch size is 144$\times$144. The number of layers $L$ in SIS and SDE modules are both set to 2.
The regularization parameters are empirically set to $\lambda_{h}=10$, $\lambda_{hl}=1$, $\lambda_{ms}=8$, and $\lambda_{rc}=3$.
All experiments were performed on a computer with a NVIDIA RTX 4090 (24 GB) GPU unless otherwise specified.

\noindent\textbf{\textbf{Evaluation metrics.}} Following the related works, the hiding performance and the hiding capacity are evaluated based on four commonly used metrics, including Peak Signal-to-Noise Ratio (PSNR), Structural Similarity Index (SSIM)~\cite{wang2004image}, Root Mean Square Error (RMSE), and Mean Absolute Error (MAE). Moreover, we employ the new Capacity-Distortion Trade-off to evaluate image hiding methods from a capacity perspective rather than from a purely distortion perspective. 

\vspace{-0.5cm}
\subsection{Evaluation Results}
\label{sec:evaluation_results}
\subsubsection{Extra-large Capacity Image Steganography}

In this section, we show both quantitative and qualitative results on extra-large capacity image hiding and recovery  with $N=9,16,25$ secret images, which have never been achieved by the other methods in the literature. 
Except our former work InvMIHNet~\cite{chen2024InvMIHNet}, we mainly compare the proposed SMILENet with ISN~\cite{Lu2021LargecapacityIS} for two reasons: i) ISN~\cite{Lu2021LargecapacityIS} is a strong baseline method and can be flexibly extended to the settings with $N > 5$ by channel-wise concatenating $N$ secret images as input; ii) ISN~\cite{Lu2021LargecapacityIS} has a relative low memory consumption among comparison methods making it feasible for comparison on the settings with $N > 5$.
We trained ISN on the task of hiding $N=$ 9 secret images on DIV2K with its default settings on a NVIDIA RTX 4090 (24G).
Meanwhile, we also trained ISN~\cite{Lu2021LargecapacityIS} for $N \geq 16$ on a NVIDIA A100 (80G) GPU since these settings lead to out of memory on the  NVIDIA RTX 4090 (24G) GPU. 

\noindent\textbf{Quantitative results.} The quantitative experimental results of the comparison methods are shown in Table ~\ref{tab:very_large_capacity}. We can consistently observe that SMILENet shows its superiority in extra-large capacity image steganography. 
Compared to our previous work, InvMIHNet~\cite{chen2024InvMIHNet}, SMILENet achieves improvements of 7.35dB in hiding and 1.55dB in recovery when hiding 9 images.
When the number of secret images increases to 25, SMILENet can still achieve 41.26dB in PSNR for hiding and  28.63dB in PSNR for recovery. 
The experimental result demonstrates that SMILENet is a strong baseline for extra-large capacity image steganography.

\noindent \textbf{Capacity-Distortion Trade-off.} Fig.~\ref{fig:capacity_rmse_SMILENet} demonstrates the Capacity-Distortion Trade-off of ISN~\cite{Lu2021LargecapacityIS}, InvMIHNet~\cite{chen2024InvMIHNet}, and SMILENet when hiding varying numbers of secret images. 
As for ISN~\cite{Lu2021LargecapacityIS}, we can see that the distortion between the cover image and the stego image measured by RMSE experiences a sharp increase as the number of secret images grows, while the hiding capacity measured by sum of normalized mutual information first slightly improves and then declines. This result indicates that the ISN method achieves to hide more secret images, but does not effectively improve the hiding capacity.
From the Capacity-Distortion Trade-off of InvMIHNet~\cite{chen2024InvMIHNet}, we can see that there is a trend in increasing of hiding capacity and distortion when hiding more secret images.
For the proposed SMILENet, We can observe that, as the number of secret images increases, the distortion remains almost the same, ranging from 2.27 to 2.45, while the hiding capacity shows a steep increase, increasing from 1.30 to 9.53. This result demonstrates that SMILENet achieves a significantly increased hiding capacity and at the same time maintains a low distortion, proving the superiority of our approach in extra-large capacity image steganography tasks.

\begin{figure*}[h]
    \centering
    \includegraphics[width=0.8\linewidth]{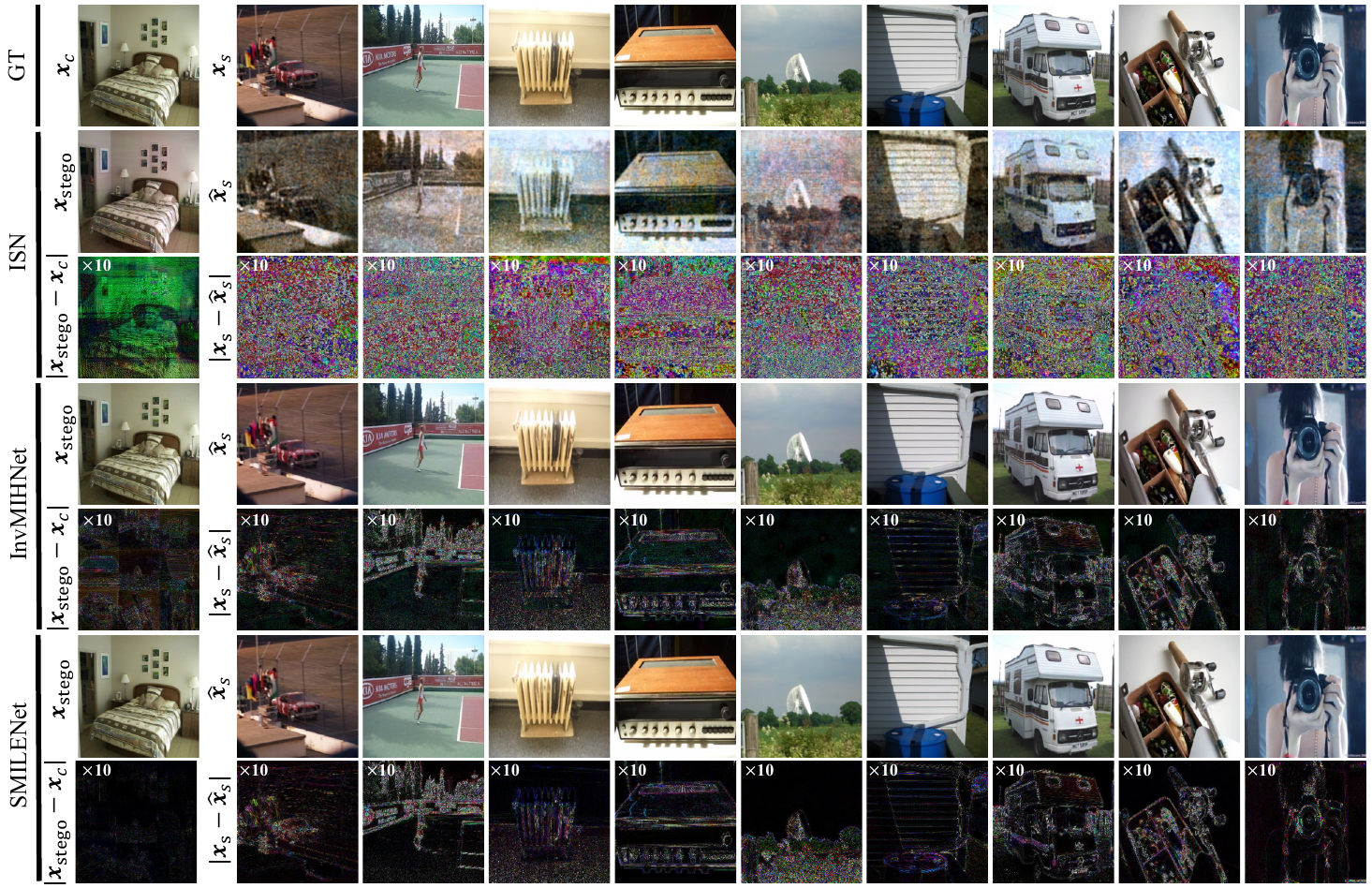}
    \caption{Visualization results of ISN~\cite{Lu2021LargecapacityIS}, InvMIHNet~\cite{chen2024InvMIHNet} and SMILENet on hiding and recovery 9 secret images evaluated on ImageNet. The $1^{\text{st}}$ column shows the ground-truth cover image, the stego image generated by ISN~\cite{Lu2021LargecapacityIS}, InvMIHNet~\cite{chen2024InvMIHNet} and SMILENet with their residual images. The remaining columns show the ground-truth secret images and the recovered secret images of the three methods with their residual images.
    }
    \label{fig:9_image_hiding}
\end{figure*}

\begin{figure*}[t]
    \centering
    \subfigure[Computational complexity]{
        \includegraphics[width=0.38\linewidth]{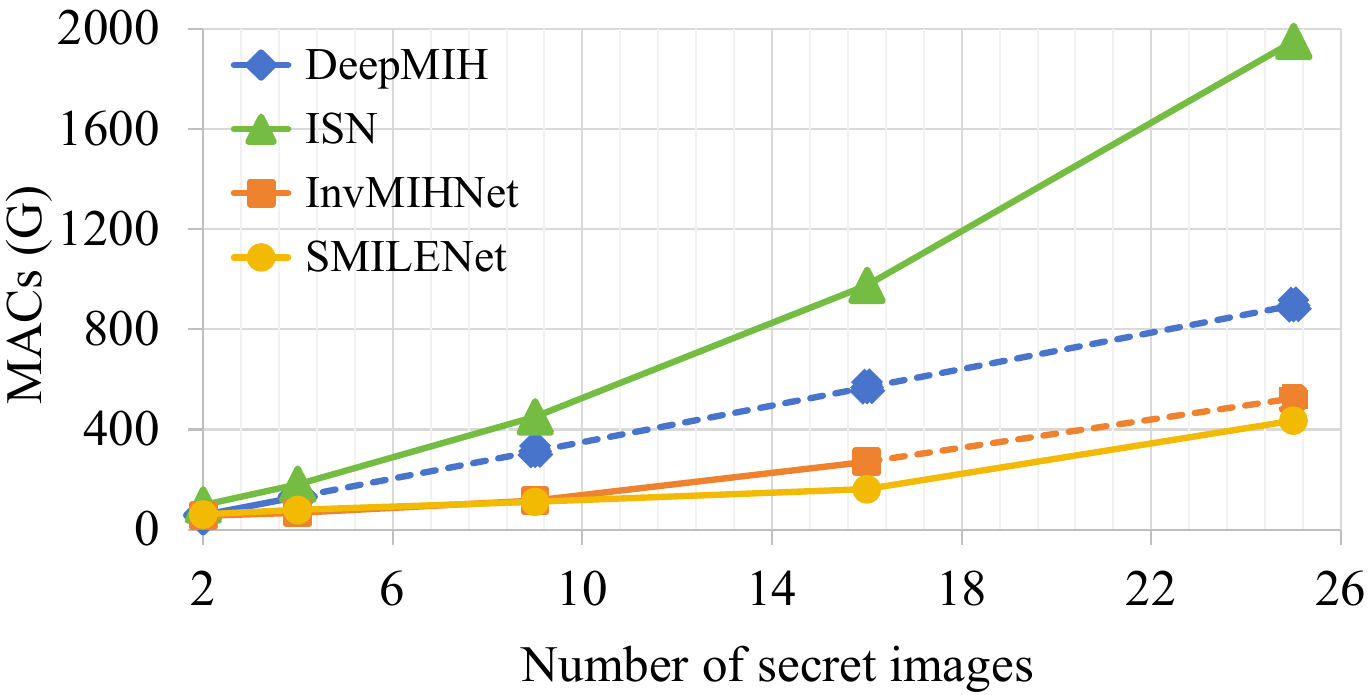} 
    }
    \hspace{0.07\textwidth}
    \subfigure[Memory consumption]{
        \includegraphics[width=0.38\linewidth]{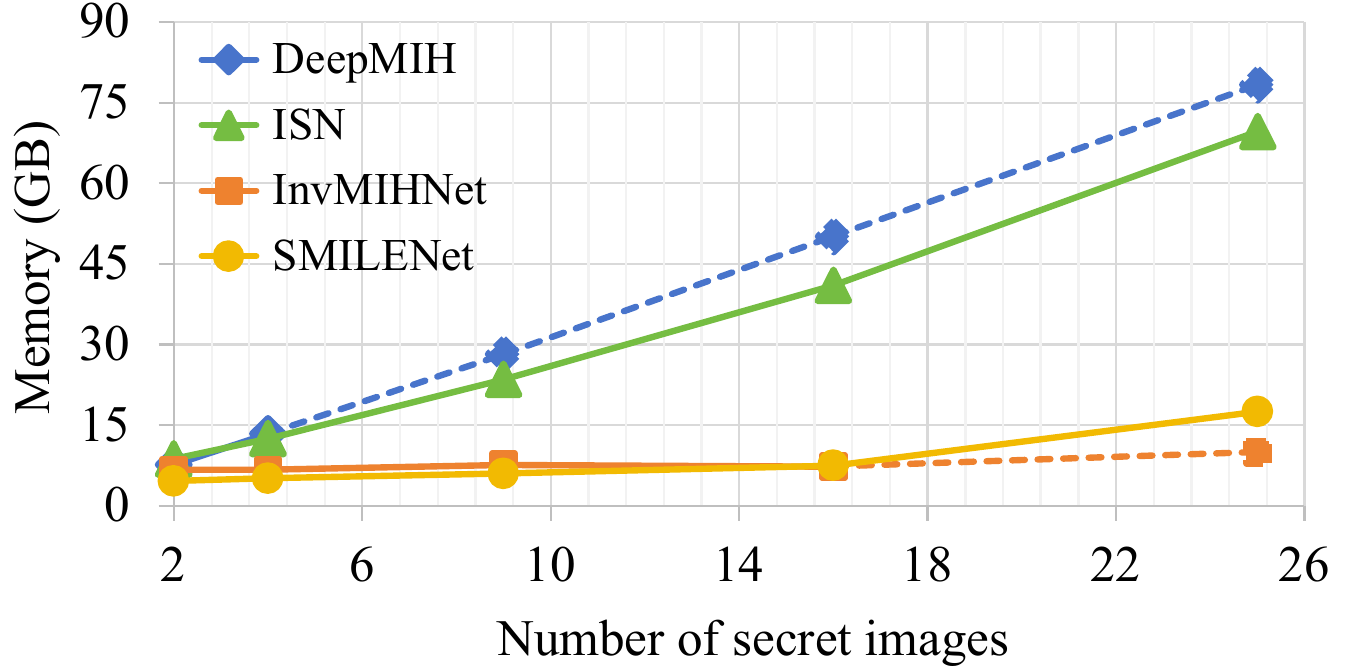} 
    }
    \caption{Comparison of different methods on the computational complexity during training and the memory consumption during testing on DIV2K dataset when the number of secret images $N \in [2,25]$. The dashed lines represent the linear interpolated values, since these models are out of the memory of our machine (RTX 4090 (24GB)).}
    \label{fig:macs_memory_nums}
    \vspace{-3pt}
\end{figure*}

\noindent\textbf{Qualitative results.} We first visualize the exemplar results of SMILENet on hiding and recovery 25 secret images, then compare the results of ISN~\cite{Lu2021LargecapacityIS}, InvMIHNet~\cite{chen2024InvMIHNet}, and SMILENet on hiding and recovery 9 secret images.

\begin{table*}[t]
\renewcommand\arraystretch{1.3}
  \centering
  \caption{The hiding and recovery results of different methods on hiding $N=4$ images evaluated on DIV2K, Paris StreetView and ImageNet. 
  (The best and the second best result in each column in bold and underlined, respectively.)}
  \scalebox{0.95}{
    \begin{tabular}{l|cccc|cccc|cccc}

    \toprule[1pt]
    \rowcolor[HTML]{EFEFEF}  \cellcolor[HTML]{EFEFEF} & \multicolumn{12}{c}{\textbf{Cover / Stego}} \\
    \cline{2-13} \rowcolor[HTML]{EFEFEF}         & \multicolumn{4}{c|}{\textbf{DIV2K} ($1024 \times 1024$)}    & \multicolumn{4}{c|}{\textbf{Paris StreetView} ($512 \times 512$)}     & \multicolumn{4}{c}{\textbf{ImageNet} ($256 \times 256$)} \\
    \cline{2-13} \rowcolor[HTML]{EFEFEF} \multirow{-3}{*}{{\textbf{Methods}}}          & \textbf{PSNR} $\uparrow$   & \textbf{SSIM} $\uparrow$  & \textbf{RMSE} $\downarrow$ & \textbf{MAE} $\downarrow$ & \textbf{PSNR} $\uparrow$   & \textbf{SSIM} $\uparrow$ & \textbf{RMSE} $\downarrow$ & \textbf{MAE} $\downarrow$  & \textbf{PSNR} $\uparrow$   & \textbf{SSIM} $\uparrow$ & \textbf{RMSE} $\downarrow$ & \textbf{MAE} $\downarrow$ \\

    \hline
    ISN~\cite{Lu2021LargecapacityIS}   & 32.90  & 0.896 & 5.84 & 4.58 &33.43&0.905&5.50&4.34 & 32.48 & 0.908 & 6.15 & 4.81 \\
    DeepMIH~\cite{guan2022deepmih}  & {33.94} & {0.904}& {5.22} & {3.78}  &{35.65}&\underline{0.948}&{4.97}& {3.30}& {34.29} & {0.924} & {5.03}& {3.71} \\
    IICNet~\cite{cheng2021iicnet} & 35.61 & 0.897 & 4.25 & 3.34 &36.13 & 0.907 & 4.01 & 3.16 & 35.18 & 0.910 & 4.48 & 3.50\\
    LF-VSN~\cite{Mou2024LFVSN} & 36.94 & 0.928 & 3.67& 2.81 & 37.49 & 0.931 & 3.50 & 2.68 &36.48 & 0.935 & 3.91 & 2.98\\
    iSCMIS~\cite{li2023iscmis} & 35.81 & 0.967 & 4.22 & 3.04  & 36.40 & 0.972 & 3.97 & 2.92 & 35.23 & 0.966 & 4.54 & 3.27\\
    InvMIHNet~\cite{chen2024InvMIHNet} &\underline{37.66} &\underline{0.945} & \underline{3.41} &\underline{2.51} & \underline{38.06} & {0.945} & \underline{3.24} & \underline{2.35} & \underline{36.86} & \underline{0.948} &\underline{3.83} & \underline{2.82}\\
    \textbf{SMILENet} & \textbf{41.10} & \textbf{0.975} & \textbf{2.29} & \textbf{1.64} &\textbf{42.25}&\textbf{0.981}&\textbf{2.01}& \textbf{1.38}&  \textbf{39.95} & \textbf{0.976} & \textbf{2.69} & \textbf{1.91} \\

    \toprule[1pt]

    \toprule[1pt]
    \rowcolor[HTML]{EFEFEF}  \cellcolor[HTML]{EFEFEF} & \multicolumn{12}{c}{\textbf{Secret / Recovery}} \\
    \cline{2-13} \rowcolor[HTML]{EFEFEF}         & \multicolumn{4}{c|}{\textbf{DIV2K} ($1024 \times 1024$)}    & \multicolumn{4}{c|}{\textbf{Paris StreetView} ($512 \times 512$)}     & \multicolumn{4}{c}{\textbf{ImageNet} ($256 \times 256$)} \\
    \cline{2-13} \rowcolor[HTML]{EFEFEF} \multirow{-3}{*}{{\textbf{Methods}}}          & \textbf{PSNR} $\uparrow$   & \textbf{SSIM} $\uparrow$  & \textbf{RMSE} $\downarrow$ & \textbf{MAE} $\downarrow$ & \textbf{PSNR} $\uparrow$   & \textbf{SSIM} $\uparrow$ & \textbf{RMSE} $\downarrow$ & \textbf{MAE} $\downarrow$  & \textbf{PSNR} $\uparrow$   & \textbf{SSIM} $\uparrow$ & \textbf{RMSE} $\downarrow$ & \textbf{MAE} $\downarrow$ \\

    \hline
    ISN~\cite{Lu2021LargecapacityIS}   & 27.65 & 0.815 & 11.31 & 7.18 &27.19&0.817&12.71&7.50 & 27.93  & 0.826& 11.37 & 7.20 \\
    DeepMIH~\cite{guan2022deepmih}  & {33.14}  & {0.930} & {5.93} & {4.16} &{35.78}&{0.944}&{4.49}& {2.98}& {32.40}  & {0.913}& {6.48} & {4.57} \\
    IICNet~\cite{cheng2021iicnet} & \underline{36.20}& \underline{0.956}& \underline{4.17}& \underline{2.86} &\underline{36.67} &\underline{0.967} & \textbf{4.01} & \underline{2.65} & \underline{35.58} & \underline{0.953} & \underline{4.54} & \underline{3.07}\\ 
    LF-VSN~\cite{Mou2024LFVSN} & 27.03 & 0.863 & 12.31 & 8.18 & 27.88 & 0.864 & 11.09 & 7.29 & 27.30 & 0.861 & 11.86 & 8.07\\
    iSCMIS~\cite{li2023iscmis} & 35.56 & 0.961 & 4.59 & 3.42 & 35.30 & 0.960 & 4.77 & 3.50 &34.70 &0.953 & 5.02 & 3.71\\
    InvMIHNet~\cite{chen2024InvMIHNet} &{33.36} &{0.939} & {6.09} &{3.79} & {33.16} & {0.946} &  {6.54} & {3.81} & {32.90} & {0.931} & {6.17} &{3.91} \\ 
    \textbf{SMILENet} & \textbf{37.11} & \textbf{0.965}& \textbf{3.96} & \textbf{2.49} 
    &\textbf{36.81}&\textbf{0.972}&\underline{4.30}& \textbf{2.46} & \textbf{36.35} & \textbf{0.963} & \textbf{4.33} & \textbf{2.73} \\
    \toprule[1pt]
    \end{tabular}%
    }
  \label{tab:conceal_four_2}%
\end{table*}%

\begin{table*}[t]
\renewcommand\arraystretch{1.3}
  \centering
  \caption{The hiding and recovery results of different methods on hiding $N=2$ images evaluated on DIV2K, Paris StreetView and ImageNet. 
  (The best and the second best result in each column is in bold and underlined, respectively.)}
  \scalebox{0.95}{
    \begin{tabular}{l|cccc|cccc|cccc}
    \toprule[1pt]
    \rowcolor[HTML]{EFEFEF}  \cellcolor[HTML]{EFEFEF} & \multicolumn{12}{c}{\textbf{Cover / Stego}} \\
\cline{2-13} \rowcolor[HTML]{EFEFEF}         & \multicolumn{4}{c|}{\textbf{DIV2K} ($1024 \times 1024$)}    & \multicolumn{4}{c|}{\textbf{Paris StreetView} ($512 \times 512$)}     & \multicolumn{4}{c}{\textbf{ImageNet} ($256 \times 256$)} \\
\cline{2-13} \rowcolor[HTML]{EFEFEF} \multirow{-3}{*}{{\textbf{Methods}}}          & \textbf{PSNR} $\uparrow$   & \textbf{SSIM} $\uparrow$  & \textbf{RMSE} $\downarrow$ & \textbf{MAE} $\downarrow$ & \textbf{PSNR} $\uparrow$   & \textbf{SSIM} $\uparrow$ & \textbf{RMSE} $\downarrow$ & \textbf{MAE} $\downarrow$  & \textbf{PSNR} $\uparrow$   & \textbf{SSIM} $\uparrow$ & \textbf{RMSE} $\downarrow$ & \textbf{MAE} $\downarrow$ \\
    \hline
    Weng \textit{et al.}~\cite{weng2019high} & 28.00 & 0.869 & 10.98 & 8.22 &29.32&0.869&9.29&7.07 & 29.79 & 0.865 & 9.04 & 6.88 \\
    Baluja~\cite{baluja2019hiding} & 31.70 & 0.890 & 6.84 & 5.27 &32.27&0.895&6.34&4.86 &31.46 & 0.887 & 6.94 & 5.29 \\
    UDH~\cite{zhang2020udh} & {37.48} & 0.940 & {3.42} & {2.82} &37.36&0.927&3.48&2.88 & 37.55 & 0.938 & 3.41 & 2.81 \\
    ISN~\cite{Lu2021LargecapacityIS}   & 37.13 & {0.975} & 3.64 & 2.72 &{40.31}&{0.966}& {2.48} &{1.89}& \underline{39.50} & {0.964} & \underline{2.78} & \underline{2.11}\\
    DeepMIH~\cite{guan2022deepmih}  & 36.99 & 0.966 & 3.84 & 2.58 &37.52&0.963&3.67&2.41 & 36.59 & 0.962 & 3.97 & 2.63 \\
    IICNet~\cite{cheng2021iicnet} & 37.07 & 0.931 & 3.62 & 2.82 & 37.51 & 0.930 & 3.43 & 2.67 &36.67 & 0.925 & 3.78 & 2.94\\
    LF-VSN~\cite{Mou2024LFVSN} & 39.52 & 0.962 & \underline{2.76} & 2.08 & 40.38 & 0.962 & \underline{2.49} & 1.88 &38.93 & 0.956 & 2.95 &2.20\\
    iSCMIS~\cite{li2023iscmis} &\underline{39.63} & \underline{0.981} & 2.82 & \underline{1.90} & \underline{40.79}& \underline{0.983}& 2.50& \underline{1.64} & 37.45 & \underline{0.969} & 3.67 & 2.49\\
    InvMIHNet~\cite{chen2024InvMIHNet} & 35.85 &0.937 & 4.25 & 3.12 & 36.64 & 0.931 & 3.86 & 2.79  & 35.26 & 0.926 & 4.51 & 3.30 \\
    \textbf{SMILENet} & \textbf{{41.53}} & \textbf{{0.982}} & \textbf{{2.27}} & \textbf{{1.59}} &\textbf{{43.25}}&\textbf{{0.986}}& \textbf{{1.82}} &\textbf{{1.24}}& \textbf{{40.30}} & \textbf{{0.976}} & \textbf{{2.60}} & \textbf{{1.81}} \\
    \toprule[1pt]

    \toprule[1pt]
     \rowcolor[HTML]{EFEFEF}  \cellcolor[HTML]{EFEFEF} & \multicolumn{12}{c}{\textbf{Secret / Recovery}} \\
\cline{2-13} \rowcolor[HTML]{EFEFEF}         & \multicolumn{4}{c|}{\textbf{DIV2K} ($1024 \times 1024$)}    & \multicolumn{4}{c|}{\textbf{Paris StreetView} ($512 \times 512$)}     & \multicolumn{4}{c}{\textbf{ImageNet} ($256 \times 256$)} \\
\cline{2-13} \rowcolor[HTML]{EFEFEF} \multirow{-3}{*}{{\textbf{Methods}}}          & \textbf{PSNR} $\uparrow$   & \textbf{SSIM} $\uparrow$  & \textbf{RMSE} $\downarrow$ & \textbf{MAE} $\downarrow$ & \textbf{PSNR} $\uparrow$   & \textbf{SSIM} $\uparrow$ & \textbf{RMSE} $\downarrow$ & \textbf{MAE} $\downarrow$  & \textbf{PSNR} $\uparrow$   & \textbf{SSIM} $\uparrow$ & \textbf{RMSE} $\downarrow$ & \textbf{MAE} $\downarrow$ \\

    \hline
    Weng \textit{et al.}~\cite{weng2019high} & 27.38 & 0.806 & 11.37 & 8.44 &27.36&0.814&11.54&8.31 & 31.03 & 0.893 & 7.53 & 5.82 \\
    Baluja~\cite{baluja2019hiding} & 28.63 & 0.849 & 9.71 & 7.30 &28.23&0.848&10.32&7.52 & 28.23 & 0.848 & 10.20 & 7.66\\
    UDH~\cite{zhang2020udh} & 25.82 & 0.727 & 15.33 & 9.81 &30.02&0.870&8.90&5.54 & 23.50 & 0.676 & 19.44 &12.34 \\
    ISN~\cite{Lu2021LargecapacityIS}& 29.43 & 0.910 & 9.24 & 6.25 &34.41&0.948&5.41&3.66 & 33.09 & 0.932 & 6.38 & 4.37 \\
    DeepMIH~\cite{guan2022deepmih}  & {38.57} & {0.973} & {3.19} & {2.17} &{37.87}&{0.975}&{3.58}& {2.29} & {37.01} & {0.966} & {3.88} & {2.62} \\
    {IICNet~\cite{cheng2021iicnet}} & \underline{40.45} & \underline{0.981} & \underline{2.54} & \underline{1.76} & \underline{42.19} & \underline{0.990} & \underline{2.09} & \underline{1.42} &38.95 & 0.976 & 3.10 & 2.12\\
    {LF-VSN~\cite{Mou2024LFVSN}} & 35.49 & 0.956 & 4.56 & 3.11 & 35.71 & 0.964 & 4.45 & 3.01 & 34.24 & 0.947 & 5.31 & 3.61\\
    {iSCMIS~\cite{li2023iscmis}} &38.82 &0.973 &3.08 &2.11 & 38.53 & 0.978 & 3.31 & 2.13 & \underline{38.96} & \underline{0.977} & \underline{3.03} &\underline{2.03}\\
    InvMIHNet~\cite{chen2024InvMIHNet} & 38.78 & 0.975 & 3.23 & 2.08 & 39.29 & 0.985 & 3.12 & 1.78 & 36.96 & 0.967 & 3.98 & 2.52\\
    \textbf{SMILENet} & \textbf{{43.25}} & \textbf{{0.990}} & \textbf{{1.90}} & \textbf{{1.20}} &\textbf{{44.64}}&\textbf{{0.995}}&\textbf{{1.68}}&\textbf{{0.96}} & \textbf{{40.94}} & \textbf{{0.983}} & \textbf{{2.57}} & \textbf{{1.63}}\\
    \toprule[1pt]
    \end{tabular}%
    }
  \label{tab:conceal_two}%
\end{table*}%

\begin{table}[!h]
\caption{ Parameters (M), computation complexity (GFlops), runtime(s) and memory consumption (GB) when testing when hiding 4 secret images of resolution $256\times256$ from ImageNet dataset.}
    \centering
     \renewcommand\arraystretch{1.27}
    \newcolumntype{C}[1]{>{\centering\let\newline\\\arraybackslash\hspace{0pt}}m{#1}}
    \begin{tabular}{l|cccc}
       \toprule[1pt]
    \rowcolor[HTML]{EFEFEF}Methods  & Params &  GFlops & Time & Memory\\
    \midrule
        ISN~\cite{Lu2021LargecapacityIS} & 1.75 & 114.66 & 0.09 & 0.6\\
        DeepMIH~\cite{guan2022deepmih} & 26.46 & 525.07 & 0.37 & 2.2 \\
        IICNet~\cite{cheng2021iicnet}& 6.49 & 301.43& 0.05 &0.9\\
        LF-VSN~\cite{Mou2024LFVSN} & 4.26 & 200.11 &0.10 & 0.7\\
        iSCMIS~\cite{li2023iscmis}&21.71 & 362.32 & 0.22 & 0.8\\
        InvMIHNet~\cite{chen2024InvMIHNet}& 5.71 & 175.33 & 0.15 & 0.8\\
        \textbf{SMILENet}& 7.79 & 238.80 & 0.21 & 0.7\\
    \toprule[1pt]
    \end{tabular}
    \label{tab:flops_4images}
    \vspace{-2pt}
\end{table}

\begin{figure}[t]
    \centering
    \includegraphics[height=4.5cm]{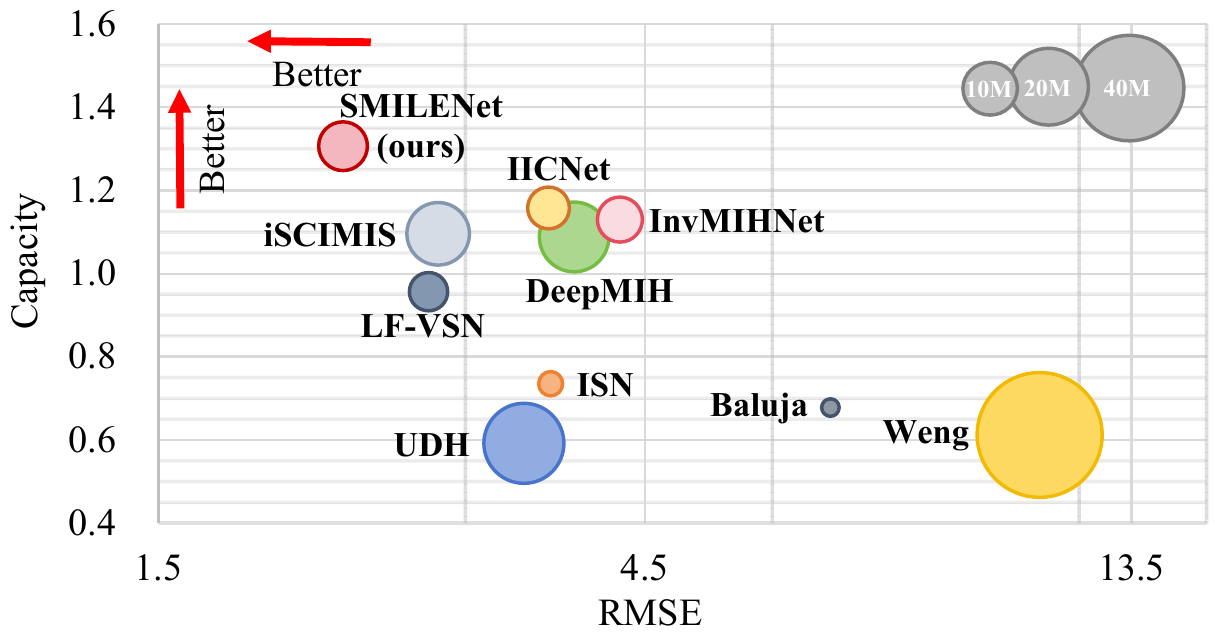}    
    \caption{The Capacity-Distortion Trade-off of different image steganography methods for hiding and recovery 2 secret images evaluated on DIV2K dataset. The results are evaluated in terms of RMSE of cover/stego pairs, hiding capacity of secret images, and the number of parameters (M). The model size is depicted as the area of the ball.}
    \label{fig:capacity_rmse}
    \vspace{-3pt}
\end{figure}

\begin{figure*}[t]
    \centering
    \includegraphics[width=0.95\linewidth]{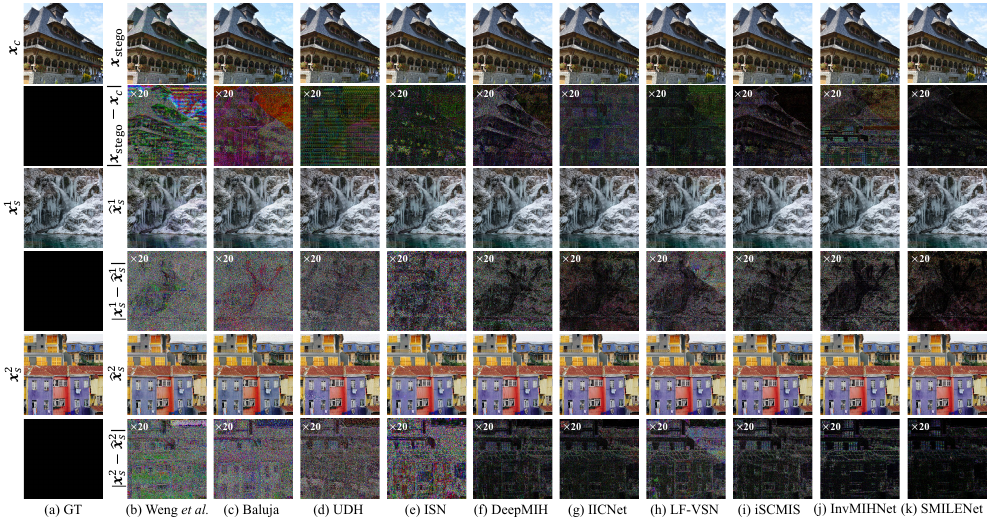}
    \caption{Visual comparisons for hiding and recovery 2 images by different approaches on DIV2K. The $1^{\text{st}}$ row and the $2^{\text{nd}}$ row denote the generated stego images and the residual images, which are enlarged 20 times for better perception. The $3^{\text{rd}}$ row and the $5^{\text{th}}$ row display the first and the second recovered secret images, with the corresponding residual images shown below each row. Better viewing in electronic version.}
    \label{fig:hiding2images}
\end{figure*}

Fig.~\ref{fig:25images} visualizes the exemplar results of SMILENet evaluated on three datasets for 25 secret images hiding and recovery. 
The $1^{\text{st}}$ - $3^{\text{rd}}$ rows show the cover image, the generated stego images and their residual image, respectively. The $4^{\text{th}}$ - $6^{\text{th}}$ rows show the original secret images, the recovered secret images and their residual images, respectively. The magnitude of all residual images are enlarged 10 times for better visualization. We can observe that i) the stego image has no visible difference compared to the cover image and their residual is too small to perceive even under 10 times magnification, ii) all secret images can be well recovered with no information interference and color distortions from other secret images. 

Fig.~\ref{fig:9_image_hiding} visualizes the exemplar results of ISN~\cite{Lu2021LargecapacityIS}, InvMIHNet~\cite{chen2024InvMIHNet} and SMILENet for 9 secret images hiding and recovery evaluated on ImageNet dataset. We can see that the stego images and recovered secret images of ISN~\cite{Lu2021LargecapacityIS} suffer from serious color distortions and their residual images looks messy. 
For InvMIHNet~\cite{chen2024InvMIHNet}, the residual between the stego image and the cover image is significantly lower than that of ISN~\cite{Lu2021LargecapacityIS}, while is still perceptible under 10 times magnification, and the recovered secret images are all well recovered with no perceptible color distortions.
In comparison, SMILENet achieves an improved hiding and recovery performance compared to InvMIHNet~\cite{chen2024InvMIHNet}. The residual between the stego image and the cover image is barely noticeable and the quality of the recovered secret images is further improved. These results demonstrates the added value of the new network architecture design and training strategy in SMILENet.

\noindent\textbf{Computational results.} 
Fig.~\ref{fig:macs_memory_nums} illustrates the computational complexity during training and the memory consumption during testing  on DIV2K dataset of ISN~\cite{Lu2021LargecapacityIS}, DeepMIH~\cite{guan2022deepmih}, InvMIHNet~\cite{chen2024InvMIHNet}, and SMILENet. We can observe a near linear relationship between the number of secret images and the computational complexity as well as the memory consumption for all comparison methods, while SMILENet is more computational and memory efficient than the other methods. 
For instance, on the task of hiding 9 secret images, the computational complexity of SMILENet is around 35.5\% and 24.7\% that of DeepMIH~\cite{guan2022deepmih} and ISN~\cite{Lu2021LargecapacityIS}, respectively. Besides, SMILENet only requires around 21.3\% and 25.5\% of memory compared to DeepMIH~\cite{guan2022deepmih} and ISN~\cite{Lu2021LargecapacityIS}, respectively. These results demonstrate efficiency of the proposed SMILENet architecture.

\subsubsection{Large Capacity Image Steganography}
In this section, we evaluate the quantitative and computational results on hiding and recovery $N=4$ secret images. The comparison methods include ISN~\cite{Lu2021LargecapacityIS}, DeepMIH~\cite{guan2022deepmih}, IICNet~\cite{cheng2021iicnet}, LF-VSN~\cite{Mou2024LFVSN}, iSCMIS~\cite{li2023iscmis} and InvMIHNet~\cite{chen2024InvMIHNet}.

\noindent\textbf{Quantitative results.} Table~\ref{tab:conceal_four_2} presents the quantitative comparison results on 4 secret images hiding and recovery results on three datasets with images of different resolutions. 
The proposed SMILENet achieves the best hiding and recovery results on all evaluation metrics.
And InvMIHNet~\cite{chen2024InvMIHNet} achieves the second best hiding results and IICNet~\cite{cheng2021iicnet} achieves the second best recovery results. 
In terms of PSNR, SMILENet surpasses ISN~\cite{Lu2021LargecapacityIS} by 8.20dB, 8.82dB and 7.47dB on DIV2K, Paris StreetView and ImageNet datasets on secret images hiding, respectively. 
Meanwhile, SMILENet also makes an improvement on the quality of secret image recovery by 9.46dB, 9.62dB and 8.42dB compared to ISN~\cite{Lu2021LargecapacityIS} on DIV2K, Paris StreetView and ImageNet datasets, respectively. 
Compared to InvMIHNet~\cite{chen2024InvMIHNet}, SMILENet achieves 3.44dB, 4.19dB, and 3.09dB higher PSNR for secret images hiding, and 3.75dB, 3.65dB and 3.45dB for secret images recovery on on DIV2K, Paris StreetView and ImageNet datasets, respectively.

\noindent\textbf{Computational results.} 
Table ~\ref{tab:flops_4images} shows the number of parameters (M), computation complexity (GFlops), runtime (s) and memory consumption (GB) of the comparison methods and SMILENet on hiding 4 images on ImageNet dataset.
The results demonstrate that SMILENet is computationally efficient compared to other methods. It requires less memory consumption and shorter runtime to complete the hiding and recovery processes, all while maintaining low computational complexity. Compared to methods like iSCMIS~\cite{li2023iscmis} and DeepMIH~\cite{guan2022deepmih}, which introduce exclusive networks for each secret image, SMILENet significantly reduces the number of trainable parameters.
The iSCMIS~\cite{li2023iscmis} achieves the third-best performance in secret recovery but demands  approximately 2.8 times more parameters and 1.3 times higher computational complexity than SMILENet. Specifically, SMILENet requires about 238.80 GFlops to hide and recover 4 secret images of resolution $256 \times 256$ with $7.79$M learnable parameters, while iSCMIS~\cite{li2023iscmis} utilizes $21.71$M parameters and takes 362.32 GFlops per group. Moreover, SMILENet takes about $50\%$ of computation time as DeepMIH~\cite{guan2022deepmih}, highlighting the efficiency of SMILENet.

\subsubsection{Low Capacity Image Steganography}
In this section, we compare the results of SMILENet on hiding and recovery 2 images  to those of Weng's method~\cite{weng2019high}, Baluja's method~\cite{baluja2019hiding}, UDH~\cite{zhang2020udh}, ISN~\cite{Lu2021LargecapacityIS}, DeepMIH~\cite{guan2022deepmih}, IICNet~\cite{cheng2021iicnet}, LF-VSN~\cite{Mou2024LFVSN}, iSCMIS~\cite{li2023iscmis}, and InvMIHNet~\cite{chen2024InvMIHNet}.

\noindent\textbf{Quantitative results.} 
Table \ref{tab:conceal_two} presents the quantitative results for hiding and recovery performances evaluated on DIV2K, Paris StreetView and ImageNet1K datasets. SMILENet achieves significant improvements compared to other comparison methods. 
When evaluating on DIV2K dataset, SMILENet achieves 41.53dB and 43.25dB in PSNR on hiding and recovery, respectively;
while iSCMIS~\cite{li2023iscmis} achieves 39.63dB and 38.82dB, and LF-VSN~\cite{Mou2024LFVSN} achieves 39.52dB and 35.49dB on hiding and recovery, respectively.
As for the performance on Paris StreetView, SMILENet achieves the best secret image recovery quality and outperforms iSCIMIS~\cite{li2023iscmis} by 5.61dB. 
On ImageNet, SMILENet also achieves the best hiding performance and achieves a 5.68dB improvement on recovery compared to iSCIMIS~\cite{li2023iscmis}.

\noindent\textbf{Capacity-Distortion Trade-off.} Fig.~\ref{fig:capacity_rmse} illustrates the Capacity-Distortion Trade-off  of different methods for hiding 2 secret images. The performance is evaluated based on
hiding capacity of secret images in terms of normalized mutual information, RMSE of cover/stego pairs, and the number of parameters (M).
As shown in the figure, SMILENet demonstrates a notable improvement by significantly increasing the hiding capacity of the secret image while simultaneously minimizing the distortion of the stego images with relative small number of network parameters, striking an effective balance between capacity and distortion.

\noindent\textbf{Qualitative results.} 
Fig.~\ref{fig:hiding2images} illustrates the exemplar visualization results of 2 images hiding and recovery of different comparison methods. We can see that all other methods suffer from clear information interference and color distortions on both the recovered secret images. In contrast, SMILENet effectively achieves to generate more imperceptible stego images and the recovered secret images are with less undesired residuals, highlighting the effectiveness of SMILENet method.

\begin{figure*}[t]
    \centering
    \subfigure[SRNet]{
         \includegraphics[width=0.38\linewidth]{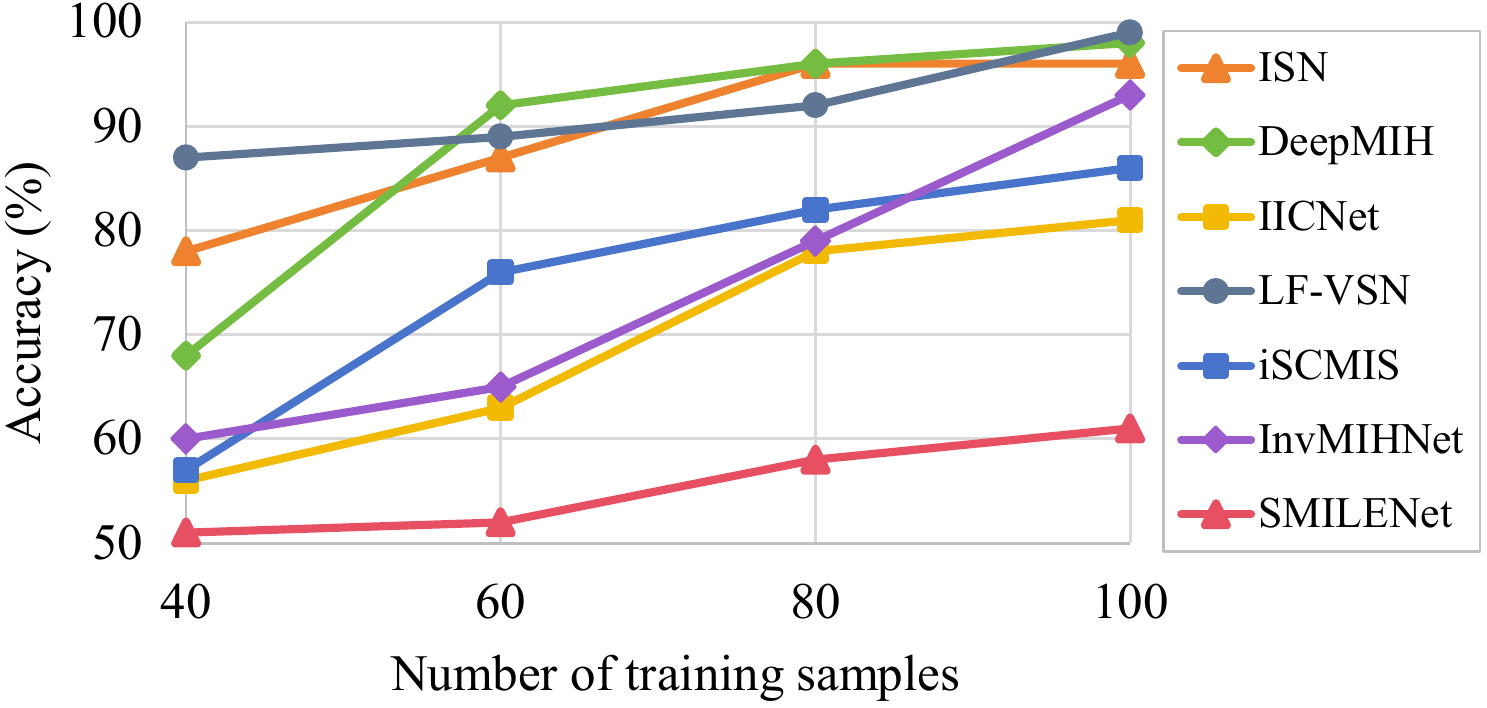} 
    }
    \hspace{0.07\textwidth}
    \subfigure[Zhu-Net]{
        \includegraphics[width=0.38\linewidth]{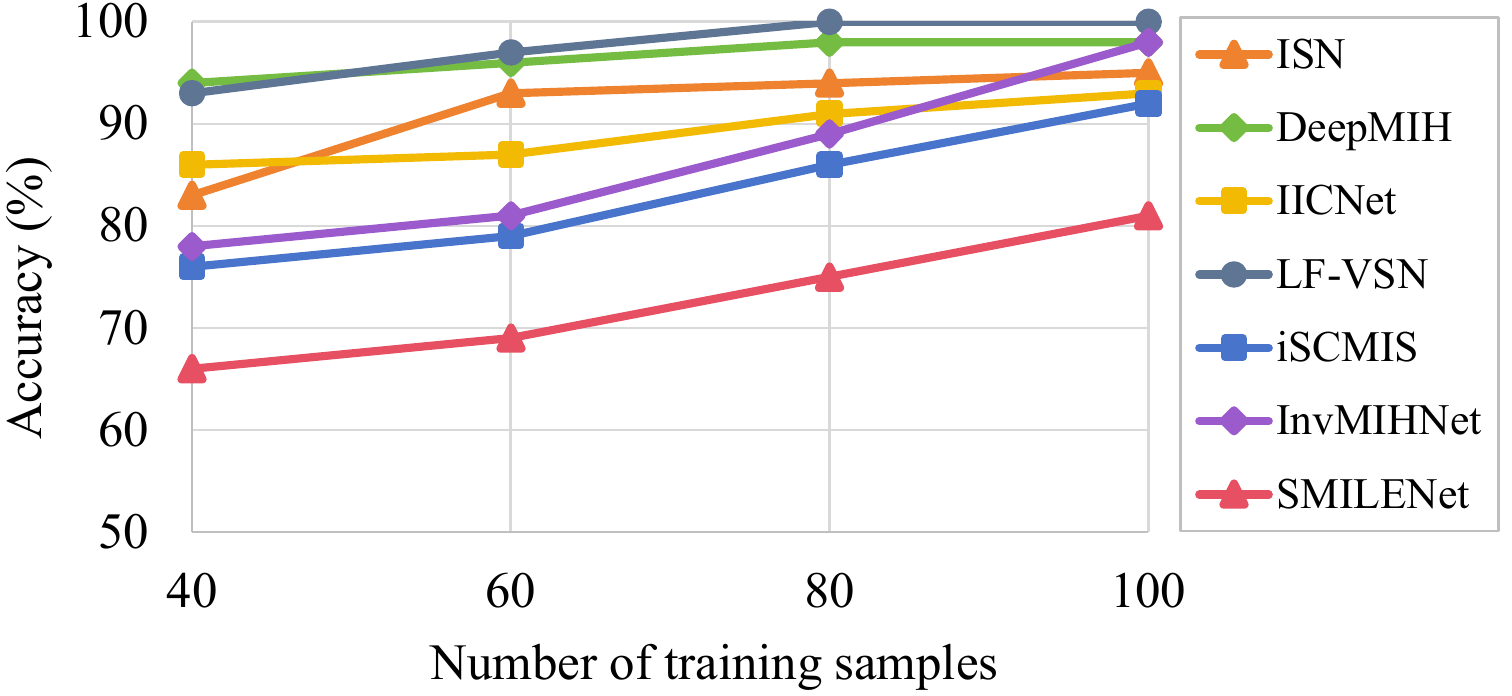} 
    }
    \caption{Steganography analysis different methods on hiding 4 secret images evaluated on ImageNet. (a) and (b) is the classification accuracy with different numbers of training samples (the lower the better) using SRNet~\cite{boroumand2018deep} and Zhu-Net~\cite{zhang2019depth}, respectively.}
    \label{fig:roc_all}
\end{figure*}

\begin{figure*}[t]
    \centering
    \includegraphics[width=1\linewidth]{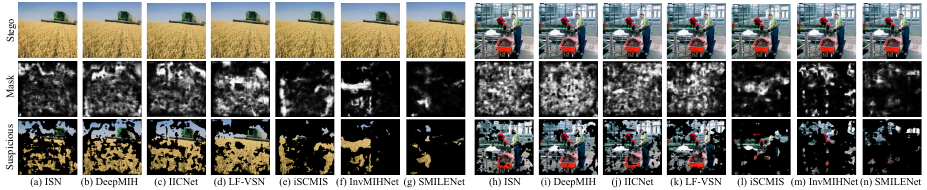}
    \caption{Forgery detection results of ManTra-Net~\cite{wu2019mantra} evaluated on the stego images generated by different methods evaluated on ImageNet. The predicted forgery mask and detected suspicious regions are displayed in the $2^{\text{nd}}$ and the $3^{\text{rd}}$ row, respectively.}
    \label{fig:mantranet}
\end{figure*}

\begin{figure*}[t]
    \centering
    \subfigure[Steganalysis with Zhu-Net]{
        \includegraphics[height=3.5cm]{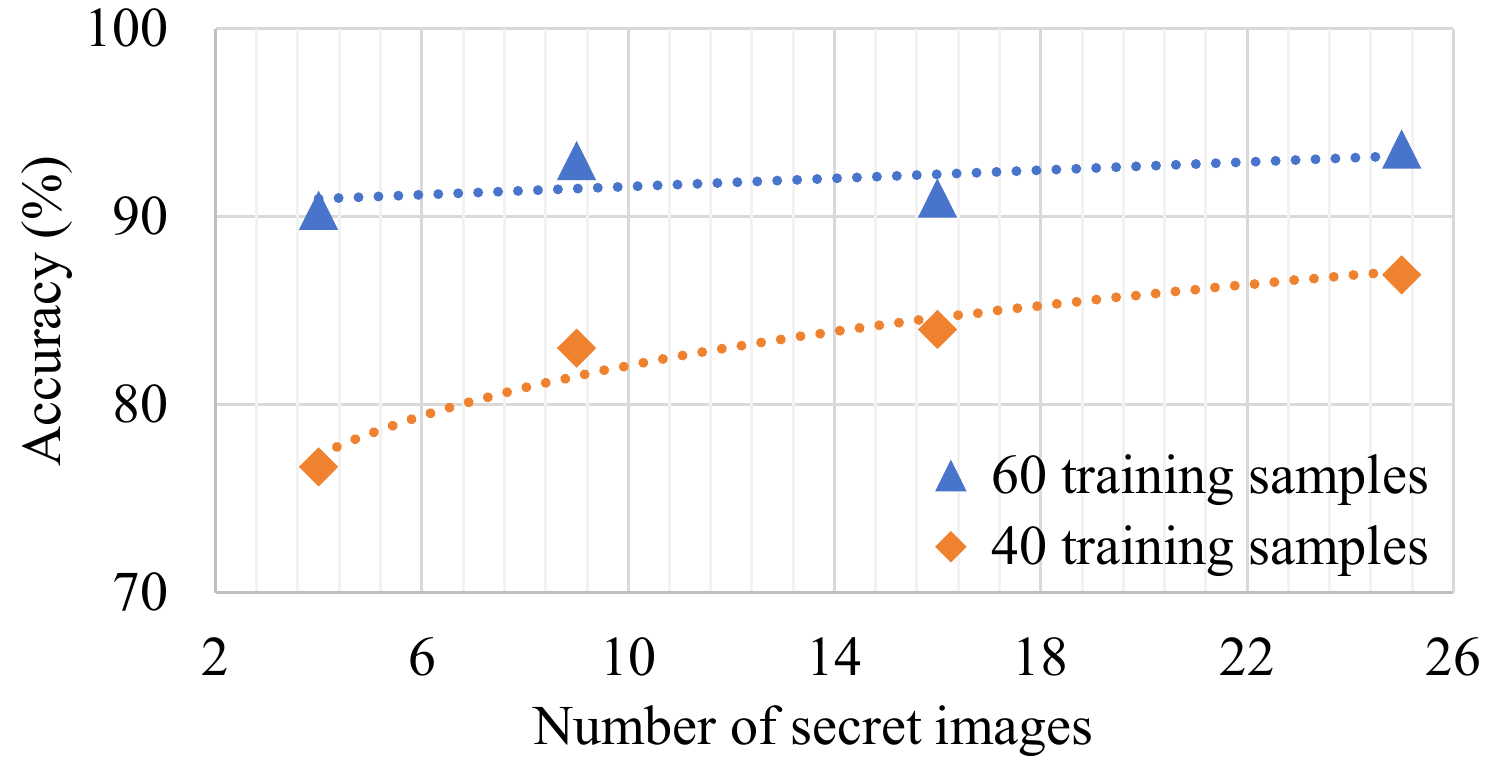} 
    }
    \hspace{0.05\textwidth}
    \subfigure[Steganalysis with ManTra-Net]{
        \includegraphics[height=3.5cm]{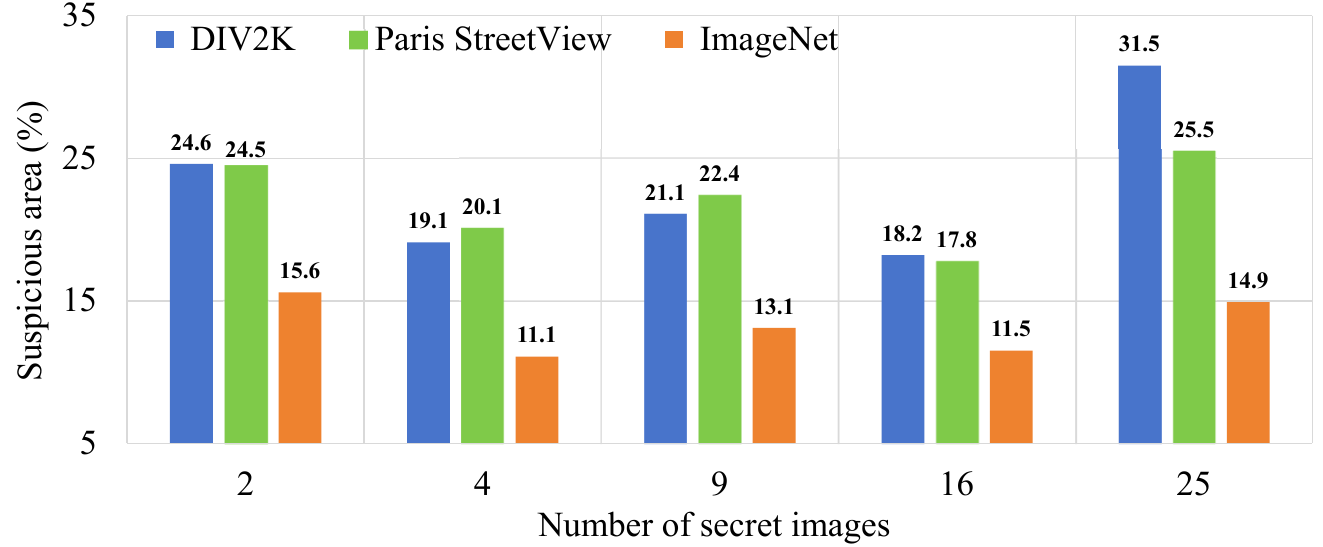} 
    }
    \caption{Steganography analysis of the stego images generated by SMILENet on extra-large capacity image steganography, including (a) the classification accuracy tested on Zhu-Net~\cite{zhang2019depth} using Paris StreetView dataset and (b) the 
    percentage of suspicious areas of stego images detected by ManTra-Net~\cite{wu2019mantra} evaluated on DIV2K, Paris StreetView, and ImageNet.}
    \label{fig:mantranet_area}
\end{figure*}

\subsubsection{Steganography Analysis}
For fair comparison, we follow the settings in ISN~\cite{Lu2021LargecapacityIS}, DeepMIH~\cite{guan2022deepmih} and iSCIMIS~\cite{li2023iscmis} which utilize three steganalysis methods including SRNet~\cite{boroumand2018deep}, Zhu-Net~\cite{zhang2019depth}, and ManTra-Net~\cite{wu2019mantra} to test anti-steganalysis ability of stego images generated by different methods including ISN~\cite{Lu2021LargecapacityIS}, DeepMIH~\cite{guan2022deepmih},  IICNet~\cite{cheng2021iicnet}, LF-VSN~\cite{Mou2024LFVSN},  iSCMIS~\cite{li2023iscmis}, InvMIHNet~\cite{chen2024InvMIHNet}, and SMILENet.
The anti-steganalysis ability is evaluated on the case of hiding 4 secret images using stego images from ImageNet and DIV2K datasets.

\noindent\textbf{Steganalysis with SRNet~\cite{boroumand2018deep} and Zhu-Net~\cite{zhang2019depth}.}
The classification accuracy with the increasing numbers of training data is shown in Fig.~\ref{fig:roc_all} (a) and (b). 
We gradually increase the number of training samples from 40 to 100 and record the detection results using 50 testing images from ImageNet. The other settings are the same to that of DeepMIH~\cite{guan2022deepmih}. 
As given in Fig.~\ref{fig:roc_all} (a), SMILENet achieves the lowest detection accuracy as the number of training samples varies from 40 to 100. 
Fig.~\ref{fig:roc_all} (b) shows the classification accuracy of Zhu-Net~\cite{zhang2019depth} for steganalysis. Compared to SRNet~\cite{boroumand2018deep}, Zhu-Net~\cite{zhang2019depth} has a stronger discriminative power, but SMILENet still has the best anti-steganalysis ability among all comparison methods.

\noindent\textbf{Steganalysis with ManTra-Net~\cite{wu2019mantra}.}
ManTra-Net~\cite{wu2019mantra} is used to detect and locate anomalous features of the generated stego images.
The steganalysis results using ManTra-Net~\cite{wu2019mantra} are given in Fig.~\ref{fig:mantranet}.
The $1^{\text{st}}$ row shows the stego images generated by different methods.
The predicted forgery mask and detected suspicious regions are displayed in the $2^{\text{nd}}$ row and the $3^{\text{rd}}$ row, respectively.
We can observe that the stego images of other six comparison method are suspicious on a large image region due to the superposition of secret contents from multiple secret images, making it easy to detect and locate suspicious contents, especially on the smooth regions. In contrast, SMILENet demonstrates better anti-steganalysis ability since most areas are not detected to be suspicious and this effectively reduces the region that can be traced by ManTra-Net~\cite{wu2019mantra}.

\begin{table}[t]
    \centering
    \caption{The percentage of suspicious pixels detected by ManTra-Net evaluated on DIV2K dataset.}
    \renewcommand\arraystretch{1.27}
    \newcolumntype{C}[1]{>{\centering\let\newline\\\arraybackslash\hspace{0pt}}m{#1}}
    \begin{tabular}{lC{1.7cm}|lC{1.7cm}}
    \toprule[1pt]
    \rowcolor[HTML]{EFEFEF}   Methods  & Suspicious &Methods & Suspicious \\
    \midrule
    ISN~\cite{Lu2021LargecapacityIS} &60.4\% & DeepMIH~\cite{guan2022deepmih} &45.2\%\\
    IICNet~\cite{cheng2021iicnet} &68.4\% &LF-VSN~\cite{Mou2024LFVSN} &42.8\%\\
    iSCMIS~\cite{li2023iscmis} &24.7\% & InvMIHNet~\cite{chen2024InvMIHNet} & 28.2\%\\
    \textbf{SMILENet} & \textbf{19.1}\%\\
    \toprule[1pt]
    \end{tabular}
    \label{tab:mantra-net_per}
    \vspace{-3mm}
\end{table}

Furthermore, we counted the percentage of suspicious pixels of the stego images detected by ManTra-Net~\cite{wu2019mantra}. The proposed SMILENet achieves a substantially lower percentage of suspicious pixels than all other comparison methods.
For instance, 
for the task of 4 secret image hiding evaluated on DIV2K dataset, the average percentage of suspicious pixels of the stego images generated by the comparison methods and SMILENet are shown in Table.~\ref{tab:mantra-net_per}. Specifically, the suspicious regions identified in IICNet~\cite{cheng2021iicnet} and ISN~\cite{Lu2021LargecapacityIS} account for 68.4\% and 60.4\% of the pixels, respectively. In contrast, our proposed SMILENet significantly reduces the proportion of suspicious pixels to only 19.9\%.
Therefore we can conclude that SMILENet has an effectively better anti-steganalysis capability than the comparison methods.

\noindent\textbf{Steganalysis on extra-large capacity setting.}
We further tested the anti-steganalysis ability of SMILENet under the extra-large capacity setting.
Fig.~\ref{fig:mantranet_area} (a) shows the classification accuracy of Zhu-Net~\cite{zhang2019depth} on Pairs StreetView and Fig.~\ref{fig:mantranet_area} (b) shows the percentage of suspicious regions detected by ManTra-Net~\cite{wu2019mantra} when stego images are hidden with different number of secret images.
In general, as the number of secret images hidden in the cover image increases, the classification accuracy of Zhu-Net~\cite{zhang2019depth} slightly improves. This can be attributed to the increased hiding capacity, which marginally facilitates distinguishing between the cover images and stego images.
Moreover, from Fig.~\ref{fig:mantranet_area} (b), when hiding different numbers of secret images, the majority of regions in the stego images generated by SMILENet consistently remain inconspicuous on three datasets. From these results, SMILENet has demonstrated its improved security against steganalysis methods, particularly in the domain of extra-large capacity image steganography.

\subsection{Ablation Study}
\label{sec:abaltion_study}

\subsubsection{Key Components of SMILENet}
Table \ref{tab:ablation_key_module} shows the ablation study on the Secret Information Selection (SIS) module, Secret Detail Enhancement (SDE) module and the conditional invertible architecture of the Invertible Cover-Driven Mosaic (ICDM) module. 
Without cover guidance, SIS and SDE modules, the performances of hiding and recovery drop by 1.92dB and 1.12dB, respectively. By incorporating the SIS module and the SDE module, the hiding and recovery performances improve by 0.85dB and 0.61dB, respectively. Removing either of these two modules leads to simultaneous performance degradation, highlighting their critical contributions to achieving extra-large capacity image steganography. Furthermore, adding guidance of cover images enhances the performances by an additional 0.44dB for hiding and 0.51dB for recovery, emphasizing the essential role of the guidance of cover image in the hiding process. 

\begin{table}[t]
  \centering
    \caption{Ablation study on the components of SMILENet when hiding 4 secret images evaluated on DIV2K.}
    \renewcommand\arraystretch{1.27}
    \begin{tabular}{ccc|cc}
    \toprule[1pt]
    \rowcolor[HTML]{EFEFEF} {Condition} & {SIS} & {SDE} &  Cover / Stego  & Secret / Recovery\\
    \midrule
    \XSolidBrush & \XSolidBrush & \XSolidBrush & 39.81 / 0.960 & 35.99 / 0.966\\
    \XSolidBrush & \Checkmark & \Checkmark & 40.66 / 0.975 & 36.60 / 0.986\\
    \Checkmark & \XSolidBrush & \Checkmark & 40.79 / 0.974 &37.04 / 0.968 \\
    \Checkmark & \Checkmark & \XSolidBrush &40.79 / 0.974 & 36.65 / 0.966\\
    \Checkmark & \Checkmark & \Checkmark & 41.10 / 0.977 & 37.11 / 0.968 \\
    \toprule[1pt]
    \end{tabular}%
    \vspace{-3pt}
  \label{tab:ablation_key_module}%
\end{table}%

\begin{table}[t]
  \centering
  \caption{Ablation study on the loss functions when hiding 4 secret images evaluated on DIV2K.}
    \renewcommand\arraystretch{1.27}
    \begin{tabular}{ccc|cc}
    \toprule[1pt]
    \rowcolor[HTML]{EFEFEF} $\mathcal{L}_{sec}$ & $\mathcal{L}_{hide}$ & $\mathcal{L}_{aux}$   & Cover / Stego  & Secret / Recovery\\
    \midrule
   \Checkmark & \XSolidBrush & \XSolidBrush &   5.83 / 0.005 & 39.99 / 0.982\\ 
   \Checkmark & \Checkmark & \XSolidBrush & 42.86 / 0.980 & 30.92 / 0.893\\
   \Checkmark & \Checkmark & \Checkmark & 41.10 / 0.977 & 37.11 / 0.968\\
    \toprule[1pt]
    \end{tabular}%
    \vspace{-2pt}
  \label{tab:wights_loss}%
\end{table}%

\begin{table}[t]
  \centering
    \caption{Ablation study on the number of blocks of ICDM module and IMSE module evaluated on DIV2K for 4 images hiding.}
    \renewcommand\arraystretch{1.27}
    \begin{tabular}{cc|cc}
    \toprule[1pt]
    \rowcolor[HTML]{EFEFEF} ICDM  & \multicolumn{1}{c|}{IMSE} & Cover / Stego  & Secret / Recovery\\
    \midrule
    0     & 16    &  40.69 / 0.973  & 36.10 / 0.961 \\
    4     & 8     & 40.99 / 0.976 & 36.51 / 0.962\\
    4     & 16    & 40.92 / 0.976 & 36.70 / 0.966 \\
    8     & 8     & 41.12 / 0.979 & 36.66 / 0.966 \\
    8     & 16    & 41.10 / 0.977 & 37.11 / 0.968  \\
    16    & 8     & 41.00 / 0.975  & 36.95 / 0.964 \\
    16    & 16    & 40.97 / 0.974 & 36.74 / 0.964  \\
    \toprule[1pt]
    \end{tabular}%
  \label{tab:ablation_InvBlocks_layers}%
 \vspace{-2pt}
\end{table}%

\subsubsection{Loss Functions}
We also performed ablation study on the loss functions. In the total loss function $\mathcal{L}_{total}$, there are three terms, \textit{i.e.}, the secret loss $\mathcal{L}_{sec}$,  hiding loss $\mathcal{L}_{hide}$ and the auxiliary loss for IMSE module $\mathcal{L}_{aux}$.
As shown in the first row of Table \ref{tab:wights_loss}, when only using the reconstruction loss for secret images $\mathcal{L}_{sec}$, the stego images differ significantly from the cover images, whose quality deteriorates to 5.83dB in PSNR on the performance of hiding.
From the $2^{\text{nd}}$ row, we can see that the loss function imposed on cover/stego and their low-frequency sub-bands $\mathcal{L}_{hide}$ improves the hiding quality to 42.86dB with costing the performance of recovery.
Furthermore, the auxiliary loss $\mathcal{L}_{aux}$ enables the IMSE module of SMILENet embed more critical secret information while providing precise cover guidance and stabilizing the optimization process. As a result, the hiding and recovery performances are enhanced to 41.10dB and 37.11dB, respectively.

\subsubsection{Numbers of Invertible Blocks}
Invertible Blocks are key building blocks of the two Invertible Image Modules. 
Table \ref{tab:ablation_InvBlocks_layers} shows the performance of SMILENet using different numbers of cInv blocks in the ICDM module and the Inv blocks of IMSE module on 4 secret images hiding and recovery task. 
We can observe that increasing the number of Inv blocks of IMSE module results in improved recovery quality with a slightly degraded hiding quality. 
The number of cInv blocks in ICDM module affects the quality of the learned representation of the Mosaic Secret Representation which is the input of the following IMSE module. With better learned representations, the hiding quality can be improved with a slight degradation on recovery. 
When there is no cInv blocks in ICDM module, the recovery performance is degraded to 36.10dB. On the other hand, an excessive number of cInv blocks in ICDM module affects both the quality of hiding and recovery. 
From the above results, we conclude that ICDM module plays an important role in both hiding and recovery processes. By considering the trade-off between performance and complexity, the default numbers of cInv blocks and Inv blocks in ICDM and IMSE modules is set to 8 and 16, respectively.

\subsubsection{Number of Channels in Invertible Blocks}

In SMILENet, the number of channels for invertible blocks in the ICDM and IMSE modules are set to 32 by default, resulting in a model size of 7.79 MB. To fully leverage the potential of SMILENet, we gradually increase the number of channels from 32 to 256.
As shown in Table ~\ref{tab:ablation_channels}, increasing the number of channels in the IMSE module doesn't yield a significant improvement in recovery performance and leads to a saturated result when expanded to 128 channels. 
Instead, the additional complexity might overfit the feature extraction, making convergence difficult during training.
On the other hand, it is evident that increasing the number of channels in ICDM module has a beneficial impact on both hiding and recovery. When the number of channel in the ICDM module is increased to 256, the hiding and recovery performances have notable improvements of 0.33dB and 0.96dB, respectively.
This can be attributed to the fact that the ICDM module is responsible for transforming a suitable form of secret representation for hiding.

\begin{table}[t]
  \centering
    \caption{Ablation study on the number of channels of ICDM module and IMSE module evaluated on DIV2K for 4 images hiding.}
    \renewcommand\arraystretch{1.27}
    \begin{tabular}{ccr|cc}
    \toprule[1pt]
    \rowcolor[HTML]{EFEFEF}ICDM & IMSE & Params& Cover / Stego  & Secret / Recovery\\
    \midrule
    32 & 32 &   7.79 & 41.10 / 0.977 & 37.11 / 0.968\\
    32 & 64 &  18.55 & 41.12 / 0.977 &36.82 / 0.967 \\
    32 & 128 & 55.27 & 41.41 / 0.976 & 25.42 / 0.914 \\
    64 & 32 &12.12 & 41.27 / 0.976  & 37.03 / 0.964\\
    128 & 32 &27.99 & 41.12 / 0.975 & 37.62 / 0.968\\
    256 & 32 & 90.31 & 41.43 / 0.976 & 38.05 / 0.970\\
    \toprule[1pt]
    \end{tabular}%
    \vspace{-2pt}
  \label{tab:ablation_channels}%
\end{table}%

\section{Conclusions}
\label{sec:conclusion}
In this paper, we introduced a novel and effective  Synergistic Mosaic InvertibLE Hiding Network (SMILENet) for extra-large capacity image stenography. 
The proposed SMILENet makes significant improvements on hiding capacity and recovery quality by hiding the Mosaic Secret Representation into the cover image which minimizes the information interference and color distortions among secret contents.  
Benefiting from the reversibility of the Invertible Cover-Driven Mosaic (ICDM) module and the Invertible Mosaic Secret Embedding (IMSE) module, the image hiding and recovery process of SMILENet are fully coupled and reversible. The Secret Information Selection (SIS) and Secret Detail Enhancement (SDE) modules further enhance the adaptiveness and controllability of SMILENet.
Meanwhile, we propose Capacity-Distortion Trade-off to evaluate the information-theoretic capacity and visual distortion.
Extensive experimental results show that the proposed SMILENet outperforms state-of-the-art methods on the imperceptibility of stego image, recovery accuracy of the secret image and security against steganalysis methods. The capability of SMILENet on extra-large capacity image steganography is further demonstrated with acceptable computational complexity and memory consumption.
In the future, it would be interesting to explore extra-large capacity video steganography by further exploiting the spatial and temporal redundancy in videos.

\bibliographystyle{IEEEtran}
\bibliography{ref}

\vfill

\end{document}